\PassOptionsToPackage{sort&compress}{natbib}
\documentclass{fairmeta}
\usepackage{enumitem}
\usepackage{needspace,wrapfig}
\usepackage{amsmath,amssymb,amsthm,mathtools}
\usepackage{tikz}
\usetikzlibrary{arrows,calc,chains,decorations.pathreplacing,fit,matrix,positioning,shapes.geometric,shapes.misc,backgrounds,patterns}
\usepackage{tabularx,float,multirow,bm,colortbl}
\usepackage{microtype}
\usepackage{pifont}

\definecolor{irisblue}{HTML}{4A90D9}
\definecolor{irispurple}{HTML}{7B61FF}
\definecolor{irisorange}{HTML}{E8853D}
\definecolor{irisgreen}{HTML}{4CAF50}
\definecolor{irisgray}{HTML}{9E9E9E}

\definecolor{rankfirst}{HTML}{98BFE6}
\definecolor{ranksecond}{HTML}{C0D8F0}
\definecolor{rankthird}{HTML}{E1EDF9}
\newcommand{\best}[1]{\cellcolor{rankfirst}#1}
\newcommand{\besttwo}[1]{\cellcolor{ranksecond}#1}
\newcommand{\bestthree}[1]{\cellcolor{rankthird}#1}

\newcommand{\cmark}{\textcolor[HTML]{228B22}{\ding{51}}}
\newcommand{\xmark}{\textcolor[HTML]{DC3545}{\ding{55}}}
\newcommand{\modelname}{AVM}

\newcommand{\secref}[1]{Section~\ref{#1}}
\newcommand{\figref}[1]{Figure~\ref{#1}}
\newcommand{\suppsecref}[1]{\secref{#1}}
\newcommand{\suppfigref}[1]{\figref{#1}}
\newcommand{\tabref}[1]{Table~\ref{#1}}

\newif\ifAVMWrapFigures

\newif\ifAVMBlurFaces
\newcommand{\AVMFaceBlurNotice}{}
\newcommand{\AVMAfterMainText}{}

\newcommand{\AVMCapabilityRowGap}{0.045in}

\newcommand{\AVMTeaser}{%
    \includegraphics[width=\textwidth]{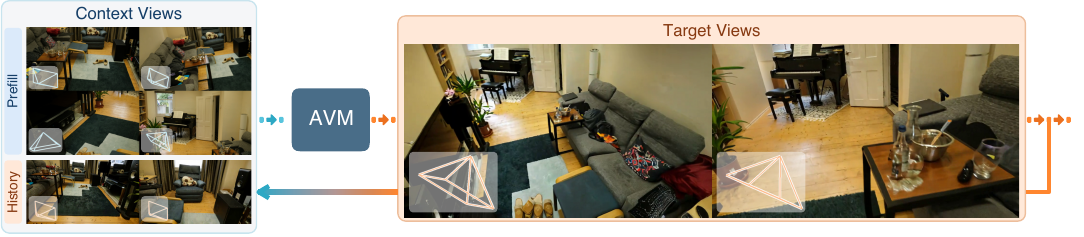}
    \captionsetup{hypcap=false}
    \captionof{figure}{\textbf{A scene as a stream of views conditioned on context.}
    AVM generates short video clips along a requested camera path with observed views as context.
    These views come from \emph{prefill} images or videos and from earlier clips, whose generated views become \emph{history} that can condition later clips.
    The same interface handles reconstruction, exploration, and revisits without fitting any 3D representation.}%
    \label{fig:teaser}
}

\makeatletter
\providecommand{\@LN}[2]{}
\providecommand{\@LN@col}[1]{}
\providecommand{\@LN@column}[1]{}
\makeatother

\renewcommand{\AVMFaceBlurNotice}{%
    \ifAVMBlurFaces\space Faces of real people are blurred for copyright reasons.\fi
}
\AVMWrapFigurestrue
\newcommand{\AVMWrapFigure}[2][r]{%
    \begin{wrapfigure}{#1}{.48\linewidth}
        \centering
        #2%
    \end{wrapfigure}%
}
\setcitestyle{numbers,square}

\title{Autoregressive View Model}
\author[1,2]{\mbox{Jensen (Jinghao) Zhou}}
\author[1]{\mbox{Samuel Rota Bulo}}
\author[1]{\mbox{Denis Rozumny}}
\author[1]{\mbox{Duncan Zauss}}
\author[1]{\mbox{Peter Hedman}}
\author[1]{\mbox{Nikhil Keetha}}
\author[2]{\mbox{Philip Torr}}
\author[2]{\mbox{Christian Rupprecht}}
\author[1,2]{\mbox{Andrea Vedaldi}}
\author[1]{\mbox{Peter Kontschieder}}
\affiliation[1]{Meta}
\affiliation[2]{University of Oxford}
\metadata[Website]{\url{https://autoregressive-view-model.github.io}}
\date{November 18, 2026}
\hypersetup{pdftitle={Autoregressive View Model}}

\abstract{We present the \underline{A}utoregressive \underline{V}iew \underline{M}odel (AVM), which generates a scene as a stream of views without constructing an explicit 3D representation.
Through a unified in-context interface, AVM conditions on images or videos to generate short clips along a requested camera path.
Visual context grounds reconstruction and interpolation, while a learned prior enables extrapolation into unseen regions.
AVM retrieves relevant views from inputs and generated history to form a bounded active context.
We fine-tune a pretrained bidirectional video diffusion transformer with flow matching in an $N$-in-1-out formulation.
We then tune a causal model on its own rollouts to improve photorealism, reduce denoising steps, and limit long-horizon drift.
We scale training to approximately 8M scenes and video clips.
We evaluate generation capacity and scene persistence across benchmarks for camera control, text control, visual quality, dynamics, sparse and 4D reconstruction, and long-horizon consistency.
The results demonstrate that AVM adapts to diverse input and output settings through one in-context interface.
}

\newcommand{\AVMBeforeReferences}{%
    \paragraph{Acknowledgment.}
    We thank Ethan John Weber for leading efforts on a predecessor internal model.
    We thank Nelson Antunes, Li Dong, Markos Georgopoulos, Xun Huang, Zhening Huang, Manuel Lopez Antequera, Norman Mueller, Lorenzo Porzi, Christian Richardt, Ayush Saraf, Johannes Schönberger, Ethan John Weber, Paul Yoo, and Zheng Zhang for helpful discussions at different stages of the project.
}

\begin{document}
\maketitle
\begin{figure}[!ht]
\centering
\AVMTeaser
\end{figure}

\begin{figure*}[p]
\centering
\vbox{\hsize=\textwidth\offinterlineskip
\hbox{\includegraphics[width=\textwidth,trim=0bp 608.4bp 0bp 0bp,clip]{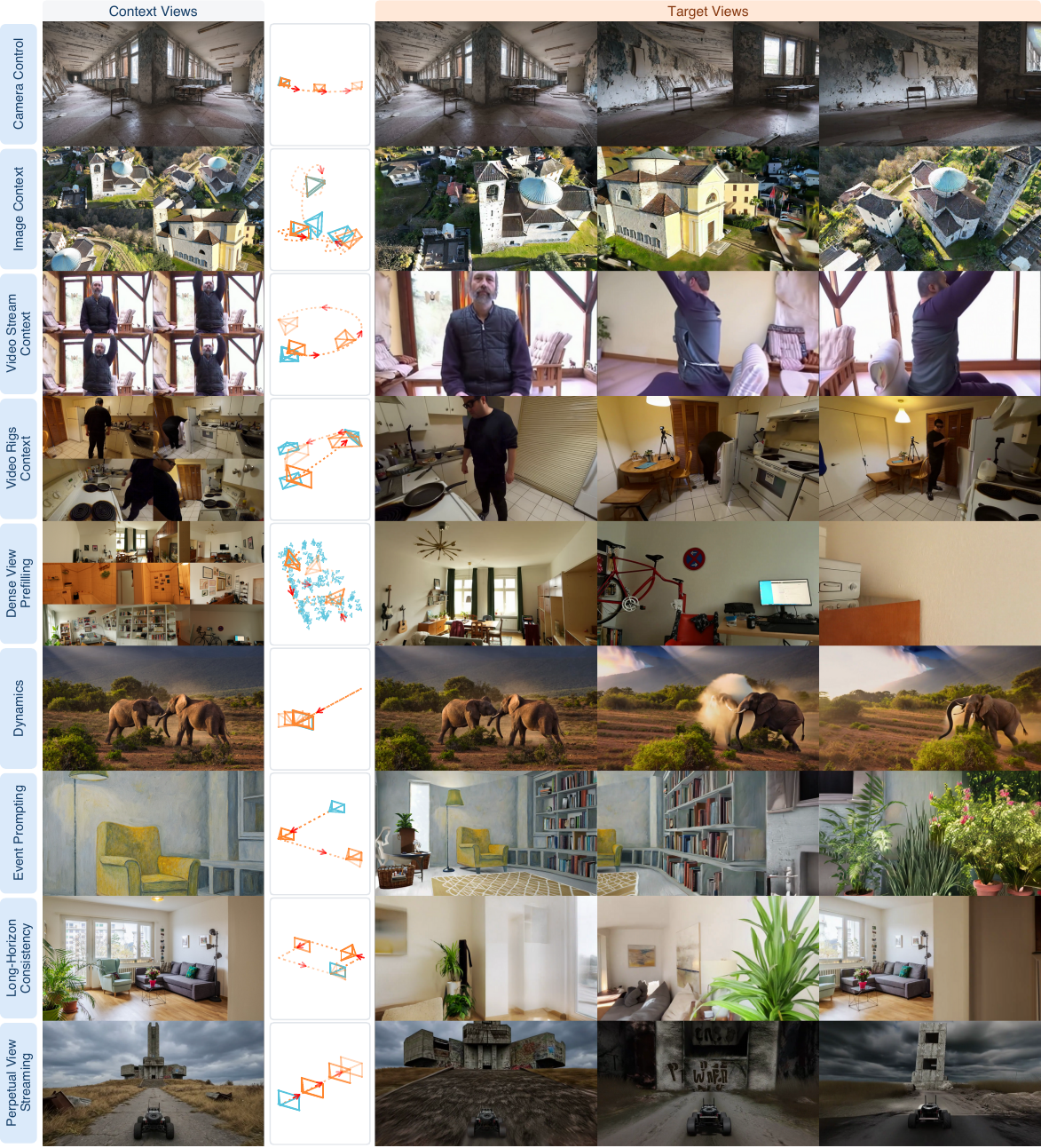}}
\hbox{\includegraphics[width=\textwidth,trim=0bp 540.9bp 0bp 12.08bp,clip]{figures/avm_capabilities.pdf}}
\vskip\AVMCapabilityRowGap
\hbox{\includegraphics[width=\textwidth,trim=0bp 473.4bp 0bp 79.58bp,clip]{figures/avm_capabilities.pdf}}
\vskip\AVMCapabilityRowGap
\hbox{\includegraphics[width=\textwidth,trim=0bp 405.9bp 0bp 147.08bp,clip]{figures/avm_capabilities.pdf}}
\vskip\AVMCapabilityRowGap
\hbox{\includegraphics[width=\textwidth,trim=0bp 338.4bp 0bp 214.58bp,clip]{figures/avm_capabilities.pdf}}
\vskip\AVMCapabilityRowGap
\hbox{\includegraphics[width=\textwidth,trim=0bp 270.9bp 0bp 282.08bp,clip]{figures/avm_capabilities.pdf}}
\vskip\AVMCapabilityRowGap
\hbox{\includegraphics[width=\textwidth,trim=0bp 203.4bp 0bp 349.58bp,clip]{figures/avm_capabilities.pdf}}
\vskip\AVMCapabilityRowGap
\hbox{\includegraphics[width=\textwidth,trim=0bp 135.9bp 0bp 417.08bp,clip]{figures/avm_capabilities.pdf}}
\vskip\AVMCapabilityRowGap
\hbox{\includegraphics[width=\textwidth,trim=0bp 68.4bp 0bp 484.58bp,clip]{figures/avm_capabilities.pdf}}
\vskip\AVMCapabilityRowGap
\hbox{\includegraphics[width=\textwidth,trim=0bp 0.9bp 0bp 552.08bp,clip]{figures/avm_capabilities.pdf}}
}
\captionsetup{font=footnotesize}
\caption{\textbf{AVM capabilities.}
The same model handles different input and output settings through a unified in-context interface.
It uses context ranging from a single image to dense image sets and synchronized video streams, supporting camera control, event prompting, and long-horizon revisits in static and dynamic scenes.
Each row shows context views (left), their camera poses and the requested trajectory (center), and target views (right).\AVMFaceBlurNotice}%
\label{fig:full-page}
\end{figure*}

\section{Introduction}%
\label{sec:intro}

We seek to generate a video of a complex 3D scene along a requested virtual camera trajectory, given one or more images or videos of the scene.
The model must reconstruct observed regions, synthesize newly visible content as the camera moves, and stay consistent with both the inputs and its own earlier outputs on revisits.
Reconstruction methods such as neural radiance fields~\citep{mildenhall2020nerf} and 3D Gaussian splatting~\citep{kerbl20233dgs} fit 3D-consistent representations to the input views and depend on dense captures for faithful rendering across viewpoints.
Video generators~\citep{brooks2024sora} can synthesize unseen content but may drift from the inputs and earlier outputs as trajectories lengthen.
Consistency becomes harder still once inputs and outputs accumulate more views than the model can condition on at once.


We present the \emph{Autoregressive View Model} (AVM), which generates a scene as a stream of views without constructing an explicit 3D representation.
Through a unified in-context interface, AVM generates short clips along a requested camera path, conditioning on posed images or videos as \emph{prefill} and on earlier generated views as \emph{history}.
Visual context grounds reconstruction and interpolation, while a pretrained video prior supports extrapolation into unseen regions.
AVM retrieves views by camera pose or timestamp from a shared pool of prefill and history, keeping each target's active context bounded even with dense inputs or long histories (\secref{sec:sampling}).
Retrieval keeps earlier views accessible beyond the local window, persistent input views anchor style and content, and recent motion context supports temporal continuity.
By varying the context, one model can reconstruct, explore, and revisit a scene.

We fine-tune a pretrained video diffusion transformer with flow matching in an \emph{$N$-in-1-out} formulation, where $N$ context slots condition one target clip.
We then derive a causal model, initialized off-policy with teacher and diffusion forcing and tuned on its own rollouts.
Distribution matching distillation and an adversarial loss improve photorealism and reduce denoising steps.
Context-augmented distillation conditions on prefill and history just as inference does, and training on longer rollouts limits drift (\secref{sec:causal}).
We train on approximately 8M scenes and video clips from real-world captures and synthetic renderings (\secref{sec:base_data}).

We evaluate generation capacity and scene persistence (\secref{sec:eval}).
From one input image, AVM achieves strong visual quality, and its camera control rivals that of view-synthesis models while substantially outperforming world models (\tabref{tab:generation_capacity}).
AVM also supports sparse reconstruction from images and 4D reconstruction from video. Retrieval improves long-horizon revisits, where AVM leads on five of six metrics (\Crefrange{tab:multi_image_video}{tab:model-ablation}).
Text alignment and dynamic degree remain limitations.
\figref{fig:full-page} shows how AVM handles dense inputs and long generated histories.

In summary, our contributions are:
1) unified in-context training for generation, reconstruction, and long-horizon consistency across base training and causal tuning,
2) unified context management and retrieval from prefill and history to support 3D consistency during generation, and
3) a unified benchmark protocol for generation capacity and scene persistence across single-image, multi-image, video, and generated history contexts.

\section{Related Work}%
\label{sec:related}
\newcommand{\singleimageicon}{\tikz[baseline={([yshift=-.5ex]current bounding box.center)},x=1ex,y=1ex,line cap=round,line join=round]{\draw[line width=.42pt] (0,0) rectangle (2,1.45); \fill (.5,1.03) circle[radius=.12ex]; \draw[line width=.42pt] (.16,.2)--(.78,.79)--(1.16,.43)--(1.55,.91)--(1.85,.2);}}
\newcommand{\imageseticon}{\tikz[baseline={([yshift=-.5ex]current bounding box.center)},x=1ex,y=1ex,line cap=round,line join=round]{\draw[line width=.42pt] (.32,.32) rectangle (2.32,1.77); \fill[white] (0,0) rectangle (2,1.45); \draw[line width=.42pt] (0,0) rectangle (2,1.45); \fill (.5,1.03) circle[radius=.12ex]; \draw[line width=.42pt] (.16,.2)--(.78,.79)--(1.16,.43)--(1.55,.91)--(1.85,.2);}}
\newcommand{\videoinputicon}{\tikz[baseline={([yshift=-.5ex]current bounding box.center)},x=1ex,y=1ex,line cap=round,line join=round]{\draw[line width=.42pt] (0,0) rectangle (2.2,1.45); \fill (.78,.26)--(.78,1.19)--(1.55,.725)--cycle;}}
\newcommand{\syncedvideoicon}{\tikz[baseline={([yshift=-.5ex]current bounding box.center)},x=1ex,y=1ex,line cap=round,line join=round]{\draw[line width=.42pt] (.32,.32) rectangle (2.52,1.77); \fill[white] (0,0) rectangle (2.2,1.45); \draw[line width=.42pt] (0,0) rectangle (2.2,1.45); \fill (.78,.26)--(.78,1.19)--(1.55,.725)--cycle;}}
\newcommand{\translationicon}{\tikz[baseline={([yshift=-.5ex]current bounding box.center)},x=1ex,y=1ex,scale=1.15,line cap=round,line join=round]{\draw[densely dotted,line width=.28pt] (.1,.12)--(1.8,.12)--(2.1,1.25)--(.4,1.25)--cycle; \draw[line width=.42pt] (.45,.34)--(.84,.9)--(1.25,.43)--(1.7,1) (1.45,.9)--(1.7,1)--(1.66,.75);}}
\newcommand{\rotationicon}{\tikz[baseline={([yshift=-.5ex]current bounding box.center)},x=1ex,y=1ex,scale=1.15,line cap=round,line join=round]{\draw[densely dotted,line width=.28pt] (1.05,.12)--(1.05,1.48); \draw[line width=.42pt] (1.75,.75) arc[start angle=0,end angle=300,x radius=.7ex,y radius=.35ex]; \draw[line width=.42pt] (1.16,.51)--(1.4,.45)--(1.22,.27);}}
\newcommand{\actioninputicon}{\tikz[baseline={([yshift=-.5ex]current bounding box.center)},x=1ex,y=1ex,line cap=round,line join=round]{\draw[line width=.35pt] (0,0) rectangle (.6,.6) (.75,0) rectangle (1.35,.6) (1.5,0) rectangle (2.1,.6) (.75,.75) rectangle (1.35,1.35); \draw[line width=.35pt] (.45,.3)--(.15,.3) (.3,.45)--(.15,.3)--(.3,.15) (1.05,.45)--(1.05,.15) (.9,.3)--(1.05,.15)--(1.2,.3) (1.65,.3)--(1.95,.3) (1.8,.45)--(1.95,.3)--(1.8,.15) (1.05,.9)--(1.05,1.2) (.9,1.05)--(1.05,1.2)--(1.2,1.05);}}
\newcommand{\geometryinputicon}{\tikz[baseline={([yshift=-.5ex]current bounding box.center)},x=1ex,y=1ex,line cap=round,line join=round]{\draw[line width=.35pt] (1.1,0)--(.1,.4)--(.1,1.2)--(1.1,1.6)--(2.1,1.2)--(2.1,.4)--cycle (.1,1.2)--(1.1,.8)--(2.1,1.2) (1.1,.8)--(1.1,0); \fill (.6,.62) circle[radius=.08ex] (1.6,.65) circle[radius=.08ex] (1.1,1.22) circle[radius=.08ex];}}
\newcommand{\statictimeicon}{\raisebox{-.4ex}{\includegraphics[height=2.2ex]{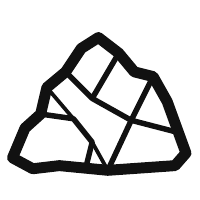}}}
\newcommand{\dynamictimeicon}{\raisebox{-.4ex}{\includegraphics[height=2.2ex]{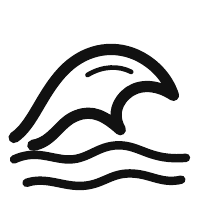}}}

\newsavebox{\singleimagecaptionbox}
\sbox{\singleimagecaptionbox}{\singleimageicon}
\newsavebox{\imagesetcaptionbox}
\sbox{\imagesetcaptionbox}{\imageseticon}
\newsavebox{\videoinputcaptionbox}
\sbox{\videoinputcaptionbox}{\videoinputicon}
\newsavebox{\syncedvideocaptionbox}
\sbox{\syncedvideocaptionbox}{\syncedvideoicon}
\newsavebox{\translationcaptionbox}
\sbox{\translationcaptionbox}{\translationicon}
\newsavebox{\rotationcaptionbox}
\sbox{\rotationcaptionbox}{\rotationicon}
\newsavebox{\actioninputcaptionbox}
\sbox{\actioninputcaptionbox}{\actioninputicon}
\newsavebox{\geometryinputcaptionbox}
\sbox{\geometryinputcaptionbox}{\geometryinputicon}
\newsavebox{\statictimecaptionbox}
\sbox{\statictimecaptionbox}{\statictimeicon}
\newsavebox{\dynamictimecaptionbox}
\sbox{\dynamictimecaptionbox}{\dynamictimeicon}

\begin{table}[t]
\centering
\caption[Comparison of related work]{\textbf{Comparison of related work.}
Context denotes images (single image \protect\usebox{\singleimagecaptionbox}, image set \protect\usebox{\imagesetcaptionbox}) or videos (single stream \protect\usebox{\videoinputcaptionbox}, synchronized multi-camera streams \protect\usebox{\syncedvideocaptionbox}).
Space denotes conditioning on action commands (\protect\usebox{\actioninputcaptionbox}), camera translation (\protect\usebox{\translationcaptionbox}), camera rotation (\protect\usebox{\rotationcaptionbox}), or point-cloud renders (\protect\usebox{\geometryinputcaptionbox}); parentheses group jointly used signals, and a slash separates alternative variants.
Time denotes static (\protect\usebox{\statictimecaptionbox}) or dynamic (\protect\usebox{\dynamictimecaptionbox}) scenes; Text indicates text-prompt conditioning; Causal indicates an available causal variant.}%
\label{tab:related_models}
\footnotesize
\setlength{\tabcolsep}{2pt}
\renewcommand{\arraystretch}{1.12}
\begin{tabular*}{\columnwidth}{@{\extracolsep{\fill}}lccccc@{}}
\toprule
\textbf{Method} & \textbf{Context} & \textbf{Space} & \textbf{Time} & \textbf{Text} & \textbf{Causal} \\
\midrule
\multicolumn{6}{l}{\textbf{View-synthesis models}} \\
GEN3C~\citep{ren2025gen3c} & \imageseticon\,\videoinputicon & \geometryinputicon & \statictimeicon\,\dynamictimeicon & \xmark & \xmark \\
Gen3R~\citep{huang2026gen3r} & \imageseticon & (\translationicon,\,\rotationicon) & \statictimeicon & \cmark & \xmark \\
SEVA~\citep{zhou2025stable} & \imageseticon & (\translationicon,\,\rotationicon) & \statictimeicon & \xmark & \xmark \\
ReCamMaster~\citep{bai2025recammaster} & \videoinputicon & (\translationicon,\,\rotationicon) & \dynamictimeicon & \cmark & \xmark \\
\midrule
\multicolumn{6}{l}{\textbf{World models}} \\
LingBot-World~\citep{robbyant2026lingbot} & \singleimageicon & (\actioninputicon\,/\,\translationicon,\,\rotationicon) & \statictimeicon\,\dynamictimeicon & \cmark & \cmark \\
HY-WorldPlay~1.5~\citep{hyworld2025} & \singleimageicon & (\actioninputicon,\,\translationicon,\,\rotationicon) & \statictimeicon\,\dynamictimeicon & \cmark & \cmark \\
DreamX-World~\citep{dreamx2026world} & \singleimageicon & (\translationicon,\,\rotationicon) & \statictimeicon\,\dynamictimeicon & \cmark & \cmark \\
\midrule
\textbf{\modelname{}} & \imageseticon\,\syncedvideoicon & (\translationicon,\,\rotationicon) & \statictimeicon\,\dynamictimeicon & \cmark & \cmark \\
\bottomrule
\end{tabular*}
\end{table}

\paragraph{View-synthesis models.}
NeRF~\citep{mildenhall2020nerf} and 3D Gaussian splatting~\citep{kerbl20233dgs} typically require dense captures and time-consuming per-scene fitting.
Feed-forward regression predicts target views directly~\citep{jin2025lvsm,jiang2025rayzer,sun2026dvsm} or infers 3D representations~\citep{yu2021pixelnerf,charatan2024pixelsplat,chen2024mvsplat}.
Diffusion models also predict views~\citep{gao2024cat3d,zhou2025stable,wu2026framecrafter,huang2026gen3r} or 3D representations~\citep{schwarz2025ggs,bahmani2026lyra}, and can additionally sample plausible unseen content.
Conditioning on renders~\citep{wu2024reconfusion,wu2025difix3d,vuong2026fixanything,delutio2026artifixer,yu2025viewcrafter} or geometric features~\citep{kang2026geonvs,lang2026roge} can improve 3D consistency.
4D novel view synthesis must preserve dynamics across viewpoints given single-camera captures~\citep{bai2025recammaster,ren2025gen3c,wang2025shapeofmotion,wu2025cat4d,vanhoorick2026anyview} or synchronized multi-camera captures~\citep{broxton2020immersive,li2022neural3dvideo,luiten2024dynamic}.
\modelname{} supports text-conditioned generation of static and dynamic scenes from posed image sets or synchronized video (\tabref{tab:related_models}).
Causal tuning enables interactive streaming with few-step denoising and key--value (KV) cache reuse across blocks.

\paragraph{World models.}
World models generate driving video, game frames, and interactive observations from inputs such as actions, text prompts, and a starting image or video prefix~\citep{hu2023gaia1,bruce2024genie,valevski2024gamenGen,huang2026solarwm,zhu2026sanawm,robbyant2026lingbot,hyworld2025}.
As rollouts grow longer, the camera can revisit locations whose earlier views lie outside the model's local context.
For scene consistency on revisits, models maintain rolling caches~\citep{robbyant2026lingbot}, condition on 3D renderings~\citep{wu2025spmem}, encode history as 3D-aware latents~\citep{yu2026worldcrafter}, or retrieve frames (by view overlap~\citep{dreamx2026world,sun2026worldplay,yu2025contextmemory}, position and yaw~\citep{chen2025interactiveworld}, or point-cloud covisibility~\citep{shen2026lyra2}).
Likewise, \modelname{} aims to preserve long-horizon scene consistency through a unified in-context interface that retrieves prefill and history views for image-based reconstruction and revisits.

\paragraph{Efficient video generation.}
Repeated denoising over long spatiotemporal sequences motivates several ways to accelerate video diffusion:
1) sparsifying attention with local windows~\citep{zhang2025slidingtile}, selected blocks~\citep{zhang2025trainablesparse}, or causal masks that block future frames or chunks and enable streaming with KV caching~\citep{teng2025magi,huang2025self,yin2025causvid} as in autoregressive language models,
2) distilling for fewer denoising steps via consistency along sampling trajectories~\citep{li2024t2vturbo,teng2025magi} or distribution matching~\citep{yin2024improved,yin2025causvid,zhu2026causal},
and 3) applying inference-specific optimizations~\citep{zhang2025turbodiffusion,liu2025teacache,zadouri2026flashattention4}.
\modelname{} starts causal tuning with teacher and diffusion forcing~\citep{chen2024diffusion}, then distills on-policy for few-step generation.

\section{Modeling}%
\label{sec:modeling}
The architecture (\secref{sec:model_design}) jointly models context latents $\mathbf z_0^{\mathrm{ctx}}$ and target latents $\mathbf z_0^{\mathrm{tgt}}$.
We first train the base model (\secref{sec:base_train}) and then apply causal tuning (\secref{sec:causal}); both stages share a unified data pipeline (\secref{sec:base_data}).

\subsection{Architecture}%
\label{sec:model_design}

\begin{figure*}[t]
\centering
\includegraphics[width=\ifAVMWrapFigures\textwidth\else.94\textwidth\fi]{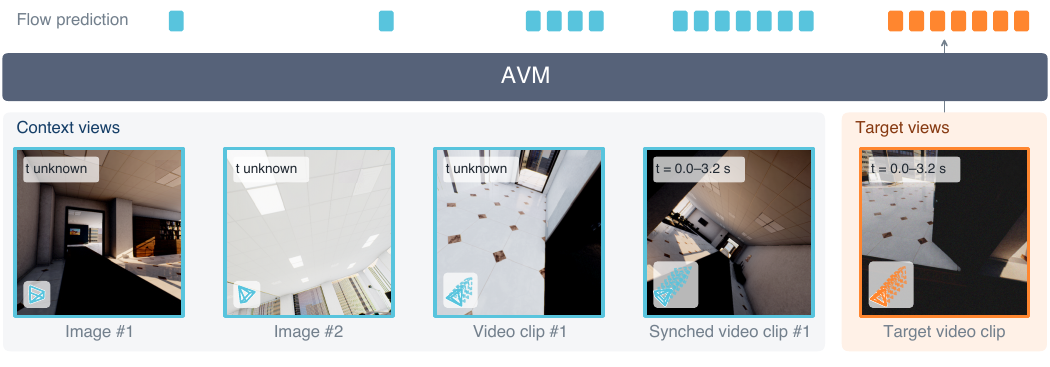}
\caption{\textbf{In-context learning.} AVM uses an \emph{$N$-in-1-out} formulation, conditioning one target video clip on $N$ context slots.
\emph{Consolidated} context (e.g., image \#1, image \#2, video clip \#1) uses views from one scene without timestamps; \emph{synced} context (e.g., synced clip \#1) provides simultaneous views from another camera at target timestamps.
The transformer processes context and target slots in one sequence.}%
\label{fig:modeling}
\end{figure*}

\paragraph{Diffusion transformer.}
\modelname{} fine-tunes a pretrained image-to-video diffusion transformer with 14 billion parameters.
A spatiotemporal variational autoencoder (VAE) maps images and videos to latents $\mathbf z_0$, with each latent representing an $8\times8\times4$ video block.
At diffusion timestep $t\in[0,1]$, the transformer ingests noisy latents $\mathbf z_t=(1-t)\mathbf z_0+t\boldsymbol\epsilon$, where $\boldsymbol\epsilon\sim\mathcal N(0,I)$.
The transformer processes the concatenated latents $[\mathbf z_t^{\mathrm{ctx}};\mathbf z_t^{\mathrm{tgt}}]$, as illustrated in \figref{fig:modeling}.
It uses FlashAttention~\citep{zadouri2026flashattention4} for full attention (as in base training and on-policy causal tuning) and FlexAttention~\citep{dong2024flexattention} for sparse attention (as in off-policy causal tuning).
Training uses BF16 mixed precision, fully sharded data parallelism (FSDP2)~\citep{pytorch2026fsdp2}, context parallelism (CP) with Ring Attention~\citep{liu2023ringattention}, and selective activation checkpointing~\citep{pytorch2025checkpointing}.

\paragraph{Conditioning.}
For each slot $i$, we concatenate the $C$-channel noisy latent $\mathbf z_{i,t}$, the $C$-channel reference latent $\mathbf c_i$, and the binary context mask $\mathbf m_i$ as $\mathbf x_{i,t}=[\mathbf z_{i,t}\Vert\mathbf c_i\Vert\mathbf m_i]$.
We set $\mathbf c^{\mathrm{ctx}}=\mathbf z_0^{\mathrm{ctx}}$ and $\mathbf m^{\mathrm{ctx}}=\mathbf 1$ for the context slots, while $\mathbf c^{\mathrm{tgt}}=\mathbf m^{\mathrm{tgt}}=\mathbf 0$ for the target slots.
This gives $2C+1$ input channels for patch embedding.
For camera pose conditioning, we use metric-scale camera poses in their original world coordinates.
For translation (\translationicon), we use rotary position embeddings (RoPE)~\citep{su2021roformer} applied to the self-attention queries and keys; for rotation (\rotationicon), we add absolute position embeddings (APE) derived from pixel-aligned camera rays to the patch embeddings.
RoPE keeps conditioning in distribution for large absolute coordinates, while ray embeddings retain orientation cues.
We add sinusoidal position embeddings (SinPE) for timestamps to the patch embeddings.
Timestamps remain constant for static scenes (\statictimeicon) and increase monotonically for dynamic scenes (\dynamictimeicon), with a randomly sampled offset during training.
T5~\citep{raffel2020t5} encodes text prompts for cross-attention.

\subsection{Training Objective}%
\label{sec:base_train}

The model predicts the flow field $\hat{\mathbf v}_{i,t}$ at each slot $i$, with ground truth $\mathbf v_{i,t}=\boldsymbol\epsilon_i-\mathbf z_{i,0}$.
The flow-matching loss~\citep{lipman2023flow} is $\mathcal L_{\mathrm{FM}}=\mathbb E\|\hat{\mathbf v}-\mathbf v\|_2^2$, with the expectation taken over slots, timesteps, and noise draws.

\paragraph{In-context learning.}
In-context learning lets language models adapt to examples without test-time parameter updates~\citep{brown2020language}.
AVM applies this principle through an \emph{$N$-in-1-out} formulation: $N$ context slots act as spatial prompts for one target video clip.
Images~\citep{zhou2025stable,gao2024cat3d} or video clips of the same scene provide \emph{consolidated} context; timestamps are discarded and temporal indices set to 0.
Camera streams at target timestamps provide \emph{synced} context~\citep{bai2025recammaster}, sharing the target views' temporal indices.
Occasionally, initial frames of the target clip provide \emph{prefix} context for start-frame or motion conditioning during chunk-wise autoregressive generation~\citep{teng2025magi} (see \secref{sec:sampling}).
Context and target views are concatenated into one sequence, as shown in \figref{fig:modeling}.

\paragraph{Error recycling.}
Error-recycling fine-tuning (ERFT)~\citep{li2026stablevideoinfinity} exposes the model to its own errors to reduce error accumulation in long chunk-wise rollouts (e.g., those used as supervision for subsequent on-policy causal distillation).
At timestep $t$, the error $\mathbf e=-t(\hat{\mathbf v}^{\mathrm{tgt}}-\mathbf v^{\mathrm{tgt}})$ is stored in a timestep-binned memory bank.
Later batches add stored errors to the noisy latents at all slots, $\tilde{\mathbf z}_t=\mathbf z_t+(1-t)\mathbf e$, and to the reference latents at context slots only, $\tilde{\mathbf c}^{\mathrm{ctx}}=\mathbf c^{\mathrm{ctx}}+\mathbf e$.
A batch receives no recycled error with probability $p$ (set to $0.5$ in practice).

\subsection{Causal Tuning}%
\label{sec:causal}
After base training, we use off-policy tuning as an initialization stage, followed by on-policy tuning on the model's own rollouts of varying length.

\paragraph{Off-policy tuning.}
We retain the flow-matching objective with ERFT from \secref{sec:base_train} and initialize causal tuning with a mixture of teacher forcing (TF)~\citep{williams1989learning} and diffusion forcing (DF)~\citep{chen2024diffusion}.
TF conditions on clean copies of the preceding blocks but halves the effective context length~\citep{huang2025self}.
DF conditions on noisy preceding blocks, creating a train--test mismatch with the denoised history used at inference~\citep{zhu2026causal}.
\modelname{} mixes TF and DF to limit this mismatch while exposing the model to imperfect context~\citep{teng2025magi}.
\emph{Prefix} context (\secref{sec:base_train}) is intrinsic to TF and can also be used with DF; \suppfigref{fig:supp-sparse-attention} illustrates the attention patterns.

\paragraph{On-policy tuning.}
We continue to tune the model on its own rollouts using distribution matching distillation (DMD)~\citep{yin2024onestep} and an adversarial loss~\citep{yin2024improved,goodfellow2014generative} to sharpen the model's output distribution and improve photorealism.
Few-step sampling with KV caching and classifier-free guidance (CFG) folding~\citep{huang2025self} enables fast rollouts.
To reduce error accumulation and encourage scene persistence over long horizons, we tune on rollouts of gradually increasing length~\citep{yang2026longlive,cui2025selfforcingpp}, with reflected and synthetic camera trajectories (\suppsecref{sec:appendix_causal_tuning}).
Distillation mirrors sampling (\secref{sec:sampling}): 1) context-augmented distillation~\citep{chen2026context,hyworld2025} conditions supervision on prefill and history views (\suppfigref{fig:supp-on-policy-distillation}), and 2) per-sample CFG distillation uses distinct guidance for static and dynamic scenes.
\tabref{tab:model-ablation} ablates per-sample CFG distillation.

\subsection{Data Modeling}%
\label{sec:base_data}

\paragraph{Data sources.}
We train on $\sim$6M labeled scenes and $\sim$2M unlabeled video clips.
Labeled scenes span indoor captures (e.g.,
ScanNet++~\citep{yeshwanth2023scannetpp} and RealEstate10K~\citep{zhou2018stereo}),
outdoor captures (e.g., DL3DV~\citep{ling2024dl3dv}), posed in-the-wild videos
(e.g., SpatialVID~\citep{wang2025spatialvid}), object videos
(e.g., uCO3D~\citep{liu2025uco3d}), synthetic environment renderings
(e.g., Aria Synthetic Environments~\citep{avetisyan2024scenescript},
ParallelDomain-4D~\citep{vanhoorick2024generative},
BlendedMVS~\citep{yao2020blendedmvs}, MVS-Synth~\citep{huang2018deepmvs},
and Spring~\citep{mehl2023spring}), game renderings, synchronized multi-camera video
(e.g., MultiCamVideo~\citep{bai2025recammaster}), and cubemap renderings
(e.g., Poly Haven~\citep{polyhaven2026}).
Labeled scenes have metric-aligned camera poses and text annotations.
We use the unlabeled, unposed in-the-wild clips for camera-unconditional training to preserve
the base video model's dynamics and text-prompt following.
Training uses five-second clips and follows a resolution curriculum, starting at $448\times448$ and then increasing to $720\times1280$.

\paragraph{Data augmentation.}
Training samples are augmented to improve model robustness.
Temporal downsampling, applied only to static-scene videos, lets fixed-length video clips span longer, more complex camera trajectories.
FLUX edits~\citep{flux2025kontext} manipulate objects or transfer style in images, encouraging coherent target views despite inconsistencies across context images.
Original and augmented inputs are randomly mixed during training.

\paragraph{Data loading.}
For each batch, a weighted sampler selects a task request specifying image count, video count, clip length, and conditioning scheme.
The request is routed to a matching dataset pool and is shared across the batch; batch size is set per request.
VAE latents for images and videos and T5~\citep{raffel2020t5} embeddings for text are precomputed offline.

\section{Sampling}%
\label{sec:sampling}
AVM samples view streams autoregressively, retrieving prefill and history views to condition target-view generation.

\paragraph{Autoregressive generation.}
Both the base and causal models can generate long videos autoregressively.
Let $L^{\mathrm{win}}=L^{\mathrm{mot}}+L^{\mathrm{tgt}}$ denote the length of a local latent window: $L^{\mathrm{mot}}$ is the length of the \emph{motion} context, which supports temporal continuity, and $L^{\mathrm{tgt}}$ is the target length.
The base model generates each target \emph{chunk} conditioned on \emph{motion} context from the previous chunk (e.g., $L^{\mathrm{mot}}=2$, $L^{\mathrm{tgt}}=19$) and reprocesses its selected context latents at every chunk.
The causal model generates shorter target \emph{blocks} within a fixed-size attention window (e.g., $L^{\mathrm{mot}}=15$, $L^{\mathrm{tgt}}=3$) and reuses past states from a KV cache.

\newcommand{\AVMContextFigure}{%
\includegraphics[width=\linewidth]{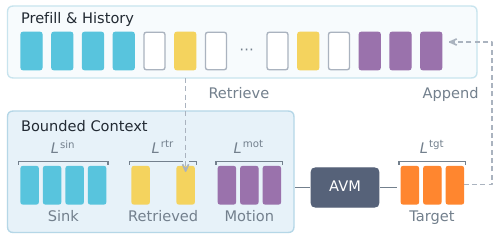}
\caption{\textbf{Context management and retrieval.}
AVM conditions target views on \emph{sink}, \emph{retrieved}, and \emph{motion} context.
Input views populate prefill; generated views join history for later retrieval.}%
\label{fig:streaming}
}
\ifAVMWrapFigures
\Needspace{.25\textheight}
\AVMWrapFigure{\AVMContextFigure}
\fi

\paragraph{Context management.}
AVM assembles a bounded context for each target, supporting diverse input and output regimes even with large prefill or history (e.g., dense view prefilling and long-horizon consistency in \figref{fig:full-page}).
As shown in \figref{fig:streaming}, \emph{sink} and \emph{retrieved} context precede a local window of recent \emph{motion} context and target views, giving a sequence length of $L=L^{\mathrm{sin}}+L^{\mathrm{rtr}}+L^{\mathrm{win}}$.
AVM retrieves views from the prefill and history pool and can retain prefill views as sink context to preserve input style and content throughout generation.
At each forward pass, AVM reassigns temporal indices~\citep{liu2026rolling,hyworld2025}: retrieved views by context type (see below), sink views at 0, and motion and target views as one consecutive segment.

During prefilling, the causal model fills its KV cache with one forward pass at zero noise per input block.
During streaming, the causal model reuses sink, retrieved, and motion KV states for all denoising steps of each target block, then runs one extra pass on the denoised block to cache its KV states for later use as motion context and, beyond the local window, for retrieval.
Retrievable KV states can be offloaded to the CPU and loaded back onto the GPU when selected.

\paragraph{Context retrieval.}
At each autoregressive step, AVM retrieves context using per-view metadata.
1) Camera poses identify spatially adjacent views as \emph{consolidated} context, with temporal indices reassigned to 0.
\suppsecref{sec:appendix_camera_retrieval} details the selection, and \suppfigref{fig:camera-retrieval-ablation} compares $L^{\mathrm{rtr}}$ across policies.
2) Timestamps identify temporally aligned views as \emph{synced} context, with temporal indices matching the corresponding targets.
3) Other retrieval cues, such as geotags on Internet images, are left for future study.
\tabref{tab:model-ablation} ablates retrieval.

\ifAVMWrapFigures\else
\begin{figure}[t]
\centering
\AVMContextFigure
\end{figure}
\fi

\begin{table*}[!t]
\centering
\caption{\textbf{Generation capacity.}
We report controllability, visual quality, and dynamics scores on DL3DV-Benchmark~\citep{ling2024dl3dv} and the static and dynamic splits of WorldScore~\citep{duan2025worldscore}.
Each model receives one starting image as visual context.
``n.a.'' denotes unavailable results.}%
\label{tab:generation_capacity}
\footnotesize
\ifAVMWrapFigures
\setlength{\tabcolsep}{0.8pt}
\else
\setlength{\tabcolsep}{1.45pt}
\fi
\begin{tabularx}{\textwidth}{@{}>{\raggedright\arraybackslash}X*{19}{c}}
\toprule
\multirow{3}{*}{Method} & Dimension & \multicolumn{8}{c}{\textbf{Controllability}} & \multicolumn{7}{c}{\textbf{Visual Quality}} & \multicolumn{3}{c}{\textbf{Dynamics}} \\
\cmidrule(lr){2-2}\cmidrule(lr){3-10}\cmidrule(lr){11-17}\cmidrule(l){18-20}
& Metric & \multicolumn{2}{c}{Camera$\uparrow$} & \multicolumn{2}{c}{Rot.$\downarrow$} & \multicolumn{2}{c}{Trans.$\downarrow$} & Obj.$\uparrow$ & Text$\uparrow$ & \multicolumn{2}{c}{Aesth.$\uparrow$} & \multicolumn{2}{c}{Image$\uparrow$} & FID$\downarrow$ & VMAF$\uparrow$ & DSim$\downarrow$ & Acc.$\uparrow$ & Deg.$\uparrow$ & Smooth.$\uparrow$ \\
\cmidrule(lr){2-2}\cmidrule(lr){3-4}\cmidrule(lr){5-6}\cmidrule(lr){7-8}\cmidrule(lr){9-9}\cmidrule(lr){10-10}\cmidrule(lr){11-12}\cmidrule(lr){13-14}\cmidrule(lr){15-15}\cmidrule(lr){16-16}\cmidrule(lr){17-17}\cmidrule(lr){18-18}\cmidrule(lr){19-19}\cmidrule(l){20-20}
& Split & D & Ws & D & Ws & D & Ws & Ws & Ws & D & Ws & D & Ws & D & D & D & Wd & Wd & Wd \\
\midrule
\multicolumn{20}{l}{\textbf{View-synthesis models}} \\
\multicolumn{2}{l}{GEN3C~\citep{ren2025gen3c}} & 68.92 & \best{83.74} & 8.22 & \best{1.55} & 0.091 & \best{0.087} & \bestthree{40.62} & 33.09 & 0.463 & 0.496 & 44.69 & 51.95 & 57.26 & 13.42 & 0.328 & n.a. & n.a. & n.a. \\
\multicolumn{2}{l}{Gen3R~\citep{huang2026gen3r}} & \bestthree{82.55} & n.a. & \bestthree{3.25} & n.a. & \bestthree{0.059} & n.a. & n.a. & n.a. & \bestthree{0.540} & n.a. & 62.84 & n.a. & \bestthree{26.33} & \bestthree{14.83} & 0.139 & n.a. & n.a. & n.a. \\
\multicolumn{2}{l}{SEVA~\citep{zhou2025stable}} & \best{97.07} & \bestthree{68.94} & \best{0.56} & \besttwo{1.66} & \best{0.011} & \bestthree{0.177} & n.a. & n.a. & 0.531 & 0.542 & 63.12 & 59.82 & \besttwo{25.04} & \besttwo{19.08} & \bestthree{0.124} & n.a. & n.a. & n.a. \\
\midrule
\multicolumn{20}{l}{\textbf{World models}} \\
\multicolumn{2}{l}{LingBot-World~\citep{robbyant2026lingbot}} & 58.45 & 3.91 & 8.17 & 36.23 & 0.099 & 0.464 & 38.75 & \best{54.07} & \besttwo{0.544} & \best{0.602} & \bestthree{63.88} & \best{65.34} & \besttwo{25.04} & 11.91 & \besttwo{0.121} & \best{68.75} & \best{35.55} & 36.59 \\
\multicolumn{2}{l}{HY-WorldPlay~1.5~\citep{hyworld2025}} & 15.86 & 3.16 & 19.23 & 18.71 & 0.283 & 0.484 & 26.25 & \bestthree{41.41} & \best{0.546} & 0.502 & \besttwo{64.70} & 53.64 & 35.71 & 12.86 & 0.151 & \bestthree{57.48} & \bestthree{26.28} & \best{76.54} \\
\multicolumn{2}{l}{DreamX-World~\citep{dreamx2026world}} & 61.73 & 2.09 & 8.83 & 23.97 & 0.107 & 0.478 & \best{42.50} & \besttwo{41.48} & 0.524 & \bestthree{0.558} & 59.48 & \bestthree{60.88} & 37.07 & 12.44 & 0.172 & 53.31 & \besttwo{27.65} & \bestthree{63.68} \\
\midrule
\multicolumn{2}{l}{\modelname{}} & \besttwo{96.48} & \besttwo{70.67} & \besttwo{0.65} & \bestthree{1.71} & \besttwo{0.013} & \besttwo{0.172} & \besttwo{41.88} & 39.65 & \besttwo{0.544} & \besttwo{0.580} & \best{65.40} & \besttwo{61.56} & \best{18.69} & \best{20.47} & \best{0.109} & \besttwo{66.36} & 24.88 & \besttwo{67.72} \\
\bottomrule
\end{tabularx}
\end{table*}

\paragraph{Classifier-free guidance.}
The base model uses 50 Euler steps with CFG\@.
Let $\hat{\mathbf v}_{+}$ and $\hat{\mathbf v}_{-}$ denote the flow predictions under full and reduced conditioning, respectively; the guided flow is $\hat{\mathbf v}_{\mathrm{CFG}}=\hat{\mathbf v}_{+}+(s-1)(\hat{\mathbf v}_{+}-\hat{\mathbf v}_{-})$.
For static scenes, $\hat{\mathbf v}_{-}$ omits context and text conditioning, with $s=1.2$ on context views and $s=5$ on target views.
For dynamic scenes, $\hat{\mathbf v}_{-}$ omits only text conditioning, with $s=5$ throughout.
The causal model uses 4 DDPM-style ancestral steps~\citep{ho2020denoising} with CFG folded in: during on-policy tuning, per-sample CFG distillation bakes each training example's static or dynamic guidance into the model (\secref{sec:causal}).

\section{Evaluation}%
\label{sec:eval}
We evaluate AVM on a wide range of generation and reconstruction tasks.
Ablations (\secref{sec:ablations}) test model and sampling choices.

\subsection{Benchmark}%
\label{sec:eval_benchmark}
\paragraph{Evaluation dimensions and metrics.}
A view model must extrapolate beyond observed views, interpolate between them, and maintain scene consistency.
We therefore evaluate two complementary dimensions:
1) Generation capacity assesses extrapolation through controllability (camera control and pose errors~\citep{ling2024dl3dv,duan2025worldscore}, object and text alignment~\citep{duan2025worldscore}), visual quality (aesthetic and imaging scores~\citep{huang2024vbench}, FID~\citep{heusel2017fid}, VMAF~\citep{netflix2016vmaf}, DreamSim~\citep{fu2023dreamsim}), and dynamics (motion accuracy~\citep{duan2025worldscore}, dynamic degree and motion smoothness~\citep{huang2024vbench});
2) Scene persistence assesses interpolation through reconstruction from prefill views and consistency on revisits to generated history, using PSNR~\citep{ling2024dl3dv}, SSIM~\citep{wang2004ssim}, LPIPS~\citep{zhang2018lpips}, place recognition (VPR) and geometric matching (SP)~\citep{dreamx2026world}, and 3D consistency (3D Cons.)~\citep{duan2025worldscore}.
See \suppsecref{sec:appendix_eval} for the protocol and \suppsecref{sec:appendix_metrics} for metric definitions.

\paragraph{Datasets, splits, and task setup.}
We detail the datasets, splits, and task settings for generation capacity in \secref{sec:eval_gencap} and for scene persistence in \secref{sec:eval_persist}.

\paragraph{Baselines.}
As discussed in \secref{sec:related} and compared in \tabref{tab:related_models}, we evaluate against view-synthesis models (GEN3C~\citep{ren2025gen3c}, Gen3R~\citep{huang2026gen3r}, SEVA~\citep{zhou2025stable}, and ReCamMaster~\citep{bai2025recammaster}) and world models (LingBot-World~\citep{robbyant2026lingbot}, HY-WorldPlay~1.5~\citep{hyworld2025}, and DreamX-World~\citep{dreamx2026world}).
Metrics use generated frames directly unless stated otherwise.

\begin{figure*}[!t]
\centering
\includegraphics[width=\textwidth]{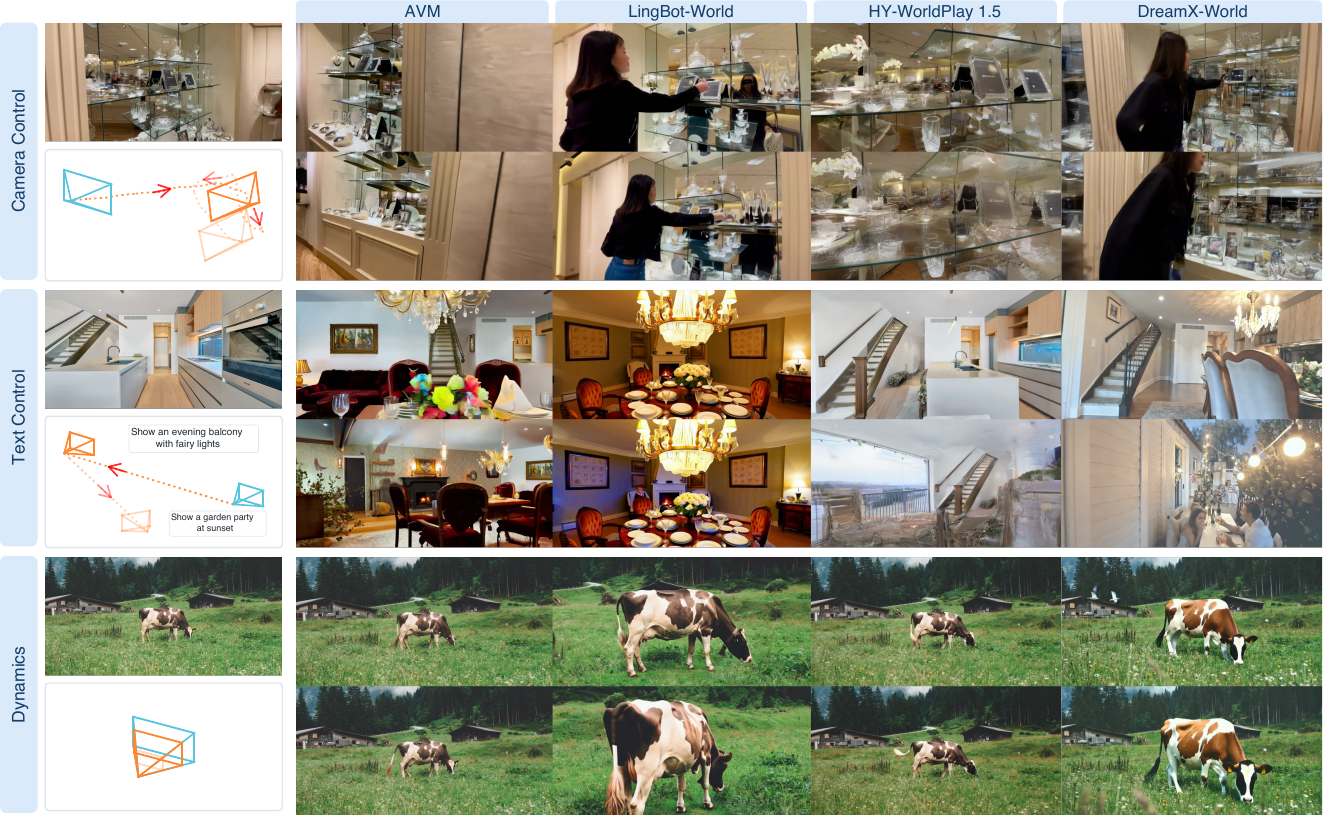}
\caption{\textbf{Qualitative comparison on generation capacity.}
We compare AVM with LingBot-World~\citep{robbyant2026lingbot}, HY-WorldPlay~1.5~\citep{hyworld2025}, and DreamX-World~\citep{dreamx2026world}.
The rows show camera control (top), text control with two prompt changes (middle), and dynamics (bottom).}%
\label{fig:qualitative-comparison}
\end{figure*}

\subsection{Generation Capacity}%
\label{sec:eval_gencap}

We compare generation capacity using a starting image as context.
For camera controllability and visual quality, we use
1) 141 DL3DV-Benchmark (D)~\citep{ling2024dl3dv} scenes with the first 81 frames of each original trajectory; and
2) 26 scenes from the static split (Ws) of WorldScore~\citep{duan2025worldscore} with the 721-frame \emph{translation \& rotation} trajectory (see \suppsecref{sec:appendix_camera_augmentation}).
For text and object controllability, we use 20 separate scenes from the static split, retaining the original 241-frame trajectory and three-segment prompts.
For dynamics, we use 32 scenes from the dynamic split (Wd) of WorldScore~\citep{duan2025worldscore}, with a static camera and 81 frames.

\paragraph{Camera controllability.}
Consistent with its view-synthesis formulation, AVM retains precise camera control during autoregressive generation, remaining competitive with view-synthesis models and substantially outperforming world models (70.67 \emph{vs.}\ the best world model's 3.91 on the Ws split).
\figref{fig:qualitative-comparison} likewise shows AVM following the camera trajectory more faithfully.

\paragraph{Text and object controllability.}
AVM is competitive in object controllability but trails all world-model baselines in text alignment (39.65 \emph{vs.}\ 54.07 for the best baseline).
In \figref{fig:qualitative-comparison}, the garden-party prompt still produces an indoor room while retaining elements of the input layout, illustrating weak text adherence alongside structural continuity.

\paragraph{Visual quality.}
AVM retains the strong visual quality of its pretrained image-to-video diffusion model (\secref{sec:model_design}), outperforming most view-synthesis models and world models across splits, notably in FID (18.69 \emph{vs.}\ the next-best 25.04 on the D split).

\paragraph{Dynamics.}
AVM retains the pretrained video model's capacity for motion, ranking second of the four models evaluated on the Wd split in both motion accuracy and smoothness.
However, its dynamic degree is the lowest (24.88 \emph{vs.}\ 35.55 for the best baseline), meaning that fewer of its videos are classified as dynamic.
The cow example in \figref{fig:qualitative-comparison} similarly preserves appearance across frames but shows only subtle pose changes.

\subsection{Scene Persistence}%
\label{sec:eval_persist}

\ifAVMWrapFigures
    \newcolumntype{V}[1]{l*{#1}{>{\centering\arraybackslash}X}}
\else
    \newcolumntype{V}[1]{>{\raggedright\arraybackslash}X*{#1}{c}}
\fi

\newcommand{\AVMImageContextTable}{%
\ifAVMWrapFigures
\begin{table}[H]
\else
\begin{table}[!b]
\fi
\centering
\caption{\textbf{Scene persistence against image context.}
Sparse reconstruction from seven posed images on DL3DV-Benchmark~\citep{ling2024dl3dv}.}%
\label{tab:multi_image_video}
\footnotesize
\setlength{\tabcolsep}{4pt}
\begin{tabularx}{\columnwidth}{@{}V{6}}
\toprule
Method & PSNR$\uparrow$ & SSIM$\uparrow$ & LPIPS$\downarrow$ & VPR$\uparrow$ & SP$\uparrow$ & 3D Cons.$\uparrow$ \\
\midrule
GEN3C~\citep{ren2025gen3c} & \bestthree{14.98} & \bestthree{0.413} & 0.441 & \bestthree{0.76} & 0.44 & \bestthree{66.77} \\
Gen3R~\citep{huang2026gen3r} & 14.30 & 0.387 & \bestthree{0.423} & \besttwo{0.90} & \bestthree{0.63} & 65.01 \\
SEVA~\citep{zhou2025stable} & \besttwo{23.91} & \besttwo{0.754} & \besttwo{0.106} & \best{0.95} & \besttwo{0.75} & \besttwo{72.01} \\
\midrule
\modelname{} & \best{24.14} & \best{0.769} & \best{0.100} & \best{0.95} & \best{0.83} & \best{73.01} \\
\bottomrule
\end{tabularx}
\end{table}
}

\newcommand{\AVMVideoContextTable}{%
\ifAVMWrapFigures
\begin{table}[H]
\else
\begin{table}[!b]
\fi
\centering
\caption{\textbf{Scene persistence against video context.}
4D reconstruction on MultiCam~\citep{bai2025recammaster} from an input video synchronized with the target views.
Mat.~Pix.\ counts are in thousands.}%
\label{tab:video_input_context}
\footnotesize
\ifAVMWrapFigures
\setlength{\tabcolsep}{4pt}
\else
\fontsize{7.6}{9.1}\selectfont
\setlength{\tabcolsep}{0.95pt}
\fi
\begin{tabularx}{\columnwidth}{@{}V{7}}
\toprule
Method & PSNR$\uparrow$ & SSIM$\uparrow$ & LPIPS$\downarrow$ & Mat.~Pix.$\uparrow$ & VPR$\uparrow$ & SP$\uparrow$ & 3D Cons.$\uparrow$ \\
\midrule
ReCamMaster~\citep{bai2025recammaster} & \besttwo{13.53} & \besttwo{0.494} & \besttwo{0.457} & \besttwo{377.79} & \besttwo{0.77} & \besttwo{0.47} & \besttwo{57.547} \\
\modelname{} & \best{21.16} & \best{0.747} & \best{0.171} & \best{450.19} & \best{0.89} & \best{0.63} & \best{64.034} \\
\bottomrule
\end{tabularx}
\end{table}
}

\newcommand{\AVMPersistenceFigure}{%
\includegraphics[width=\linewidth,height=.9\textheight,keepaspectratio]{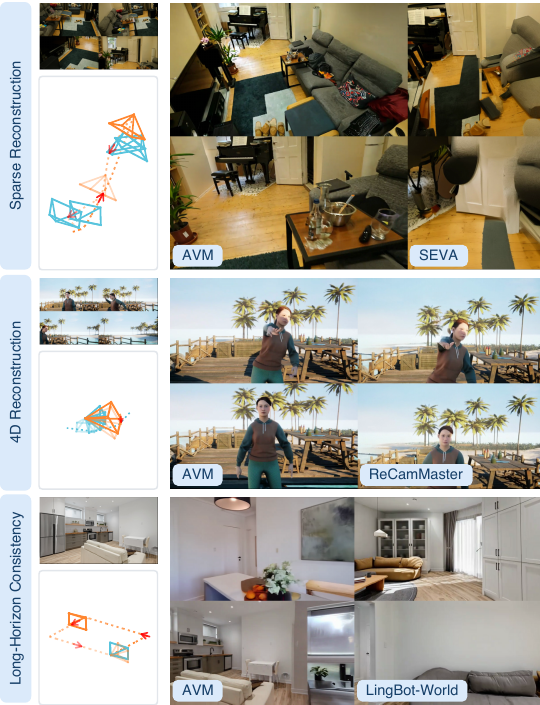}
\caption{\textbf{Qualitative comparison on scene persistence.}
We compare AVM with SEVA~\citep{zhou2025stable} for sparse reconstruction with multiple images as context (top), ReCamMaster~\citep{bai2025recammaster} for 4D reconstruction from video context (middle), and LingBot-World~\citep{robbyant2026lingbot} for revisits with long-horizon history context (bottom).}%
\label{fig:scene-persistence-comparison}
}
We compare scene persistence against different types of context.
To measure how well generated views are grounded in the input views, we evaluate on
1) the D split with the same 141 scenes, using seven input images as context; and
2) 24 MultiCam~\citep{bai2025recammaster} held-out test scenes, using the video from one camera rig as context to reconstruct 81 synchronized frames from a second camera rig.
\ifAVMWrapFigures
\par
\Needspace{.60\textheight}
\AVMWrapFigure[l]{\AVMPersistenceFigure}
\fi
To evaluate consistency with history views, we use the same 26 scenes from the Ws split with the 721-frame \emph{loop closure} trajectory (see \suppsecref{sec:appendix_camera_augmentation}).
Revisit consistency compares temporally separated generated frames, paired by proximity in camera position and yaw~\citep{dreamx2026world}.

\ifAVMWrapFigures\else
\begin{figure}[!t]
\centering
\AVMPersistenceFigure
\end{figure}
\fi

\paragraph{Persistence against image context.}
In sparse reconstruction, AVM matches or surpasses view-synthesis models on all reported metrics (\tabref{tab:multi_image_video}).
Gains over SEVA are modest in fidelity but clearer in geometric matching (0.83 \emph{vs.}\ 0.75).
The room example in \figref{fig:scene-persistence-comparison} shows clearer furniture details and more coherent room structure than SEVA, illustrating higher visual quality and greater robustness to viewpoint changes.

\ifAVMWrapFigures\else
\AVMImageContextTable
\AVMVideoContextTable
\fi

\paragraph{Persistence against video context.}
AVM uses the same interface for 4D reconstruction from video context, outperforming ReCamMaster in PSNR (21.16 \emph{vs.}\ 13.53) and all other metrics in \tabref{tab:video_input_context}.
\figref{fig:scene-persistence-comparison} also shows better preservation of character appearance and pose.

\paragraph{Persistence over history context.}
AVM retrieves generated views from outside its local window to maintain long-horizon consistency on revisits, whereas world models without explicit history retrieval cannot directly reuse those earlier views.
AVM outperforms world models in revisit PSNR (28.04 \emph{vs.}\ the best baseline's 12.29) in \tabref{tab:persistence} and preserves the kitchen layout in \figref{fig:scene-persistence-comparison}.
The ablation confirms retrieval's contribution (\secref{sec:ablations}).

\ifAVMWrapFigures
\WFclear
\begingroup
\setlength{\intextsep}{6pt plus 2pt minus 1pt}
\AVMImageContextTable
\AVMVideoContextTable
\fi

\ifAVMWrapFigures
\begin{table}[H]
\else
\begin{table}[!t]
\fi
\centering
\caption{\textbf{Scene persistence over history context.}
Revisit consistency on the Ws split of WorldScore~\citep{duan2025worldscore}.
``n.a.''\ denotes unavailable geometric matching scores.}%
\label{tab:persistence}
\footnotesize
\setlength{\tabcolsep}{1.5pt}
\begin{tabularx}{\columnwidth}{@{}V{6}}
\toprule
Method & PSNR$\uparrow$ & SSIM$\uparrow$ & LPIPS$\downarrow$ & VPR$\uparrow$ & SP$\uparrow$ & 3D Cons.$\uparrow$ \\
\midrule
LingBot-World~\citep{robbyant2026lingbot} & \bestthree{11.06} & \besttwo{0.412} & \besttwo{0.697} & \besttwo{0.49} & \besttwo{0.12} & \best{57.00} \\
HY-WorldPlay~1.5~\citep{hyworld2025} & 10.70 & \bestthree{0.411} & 0.891 & 0.13 & n.a. & 15.76 \\
DreamX-World~\citep{dreamx2026world} & \besttwo{12.29} & 0.409 & \bestthree{0.727} & \bestthree{0.40} & \bestthree{0.07} & \bestthree{47.86} \\
\midrule
\modelname{} & \best{28.04} & \best{0.836} & \best{0.136} & \best{0.92} & \best{0.77} & \besttwo{53.13} \\
\bottomrule
\end{tabularx}
\end{table}

\subsection{Ablation Studies}%
\label{sec:ablations}

We compare model variants and inference strategies in \tabref{tab:model-ablation} using the settings of \secref{sec:eval_gencap} and \secref{sec:eval_persist}.
We report the camera control score (CC) on the D split;
visual quality (VQ) with global FID on the D split;
dynamics (DY) with motion accuracy on the Wd split;
sparse reconstruction (SR) with PSNR on the D split; and
long-horizon consistency (LC) with revisit PSNR on the Ws split.

\ifAVMWrapFigures
\begin{table}[H]
\else
\begin{table}[!t]
\fi
\centering
\caption{\textbf{Ablation studies.}
Effects of model tuning, per-sample CFG distillation, retrieval, and active-window length.}%
\label{tab:model-ablation}
\footnotesize
\setlength{\tabcolsep}{2pt}
\begin{tabularx}{\columnwidth}{@{}V{5}}
\toprule
& \textbf{CC}$\uparrow$ & \textbf{VQ}$\downarrow$ & \textbf{DY}$\uparrow$ & \textbf{SR}$\uparrow$ & \textbf{LC}$\uparrow$ \\
\midrule
\modelname{} & \bestthree{96.48} & \besttwo{18.69} & \bestthree{66.36} & \besttwo{24.14} & \bestthree{28.04} \\
\midrule
\multicolumn{6}{l}{\textbf{Model variant}} \\
\modelname{} (base model) & \best{97.12} & 19.86 & \besttwo{66.98} & \best{24.32} & \best{28.53} \\
\modelname{} (causal init model) & 88.90 & 34.86 & 56.41 & 18.97 & 10.12 \\
\modelname{} (w/o per-sample CFG dist.) & 93.12 & \best{18.34} & \best{67.21} & 22.30 & \besttwo{28.12} \\
\midrule
\multicolumn{6}{l}{\textbf{Inference strategy}} \\
\modelname{} ($L^{\mathrm{rtr}}=0$) & -- & -- & -- & -- & 11.13 \\
\modelname{} ($L^{\mathrm{win}}=21$) & \besttwo{96.54} & \bestthree{18.88} & 66.25 & \bestthree{24.07} & \bestthree{28.04} \\
\modelname{} ($L^{\mathrm{win}}=15$) & 96.31 & 19.02 & 66.17 & 23.98 & 28.00 \\
\bottomrule
\end{tabularx}
\end{table}

\paragraph{Model variants.}
On-policy tuning largely restores the base model's camera control and reconstruction after causal initialization, notably in revisit PSNR (28.04 \emph{vs.}\ 10.12), which supports tuning on the model's own rollouts.
Per-sample CFG distillation improves camera control and sparse reconstruction but slightly worsens FID and motion accuracy, exposing a guidance trade-off.

\paragraph{Inference strategies.}
Disabling retrieval ($L^{\mathrm{rtr}}=0$) reduces revisit PSNR from 28.04 to 11.13.
In contrast, reducing the active window from 21 to 15 latent frames barely changes revisit PSNR (28.04 \emph{vs.}\ 28.00) or the other scores, supporting a bounded active context with retrieved history.
\ifAVMWrapFigures\endgroup\fi

\ifAVMWrapFigures\Needspace{7\baselineskip}\fi
\section{Conclusion}%
\label{sec:conclusion}
\modelname{} generates a scene as a stream of views, using one in-context interface for reconstruction, exploration, and revisiting without fitting an explicit 3D representation.
Observed views ground reconstruction and interpolation, while a pretrained video prior supports extrapolation into unseen regions.
Retrieval from prefill and generated history lets the model use dense inputs and revisit earlier locations with a bounded active context.
Results across generation and reconstruction tasks show strong camera controllability and visual quality, with retrieval improving revisit fidelity.
Future directions include adapting stronger open-weight video priors such as MiniMax H3~\citep{minimax2026h3}, adding optional geometric proxies for finer control over scene structure, and integrating AVM into agentic workflows.

\clearpage
\AVMAfterMainText
\AVMBeforeReferences

{\small
\bibliographystyle{ieeenat_fullname}
\bibliography{paper}
}

\ifAVMWrapFigures\FloatBarrier\else\clearpage\fi
\appendix
\raggedbottom
\ifAVMWrapFigures\Needspace{.60\textheight}\fi
\section{Camera Primitives}%
\label{sec:appendix_camera_augmentation}
We use the camera primitives detailed in \figref{fig:supp-camera-trajectory-augmentation} to augment camera paths during on-policy distillation (\secref{sec:appendix_causal_tuning}) and to construct evaluation trajectories (\secref{sec:eval_gencap} and \secref{sec:eval_persist}).

\ifAVMWrapFigures
\begin{figure}[H]
\else
\begin{figure*}[t]
\fi
\centering
\includegraphics[width=\linewidth]{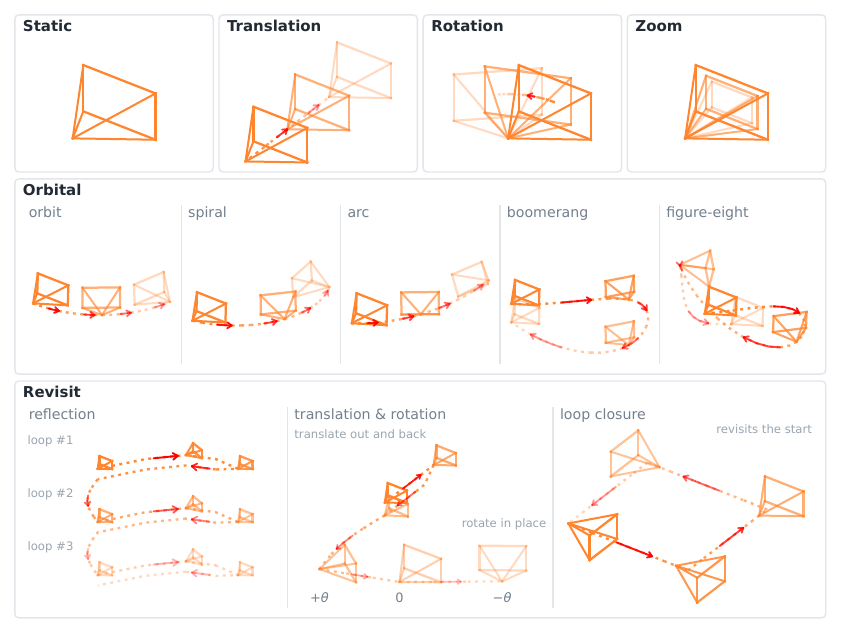}
\caption{\textbf{Camera primitives.}
Reflection repeats a continuous out-and-back path.
The translation \& rotation path returns to the start before rotating in place.
Overlapping passes and turn poses are offset for clarity.
See \tabref{tab:camera_augmentation} for the augmentation settings.}%
\label{fig:supp-camera-trajectory-augmentation}
\ifAVMWrapFigures
\end{figure}
\else
\end{figure*}
\fi

\section{Causal Tuning}%
\label{sec:appendix_causal_tuning}

\newcommand{\AVMSparseAttentionFigure}{%
\includegraphics[width=\linewidth]{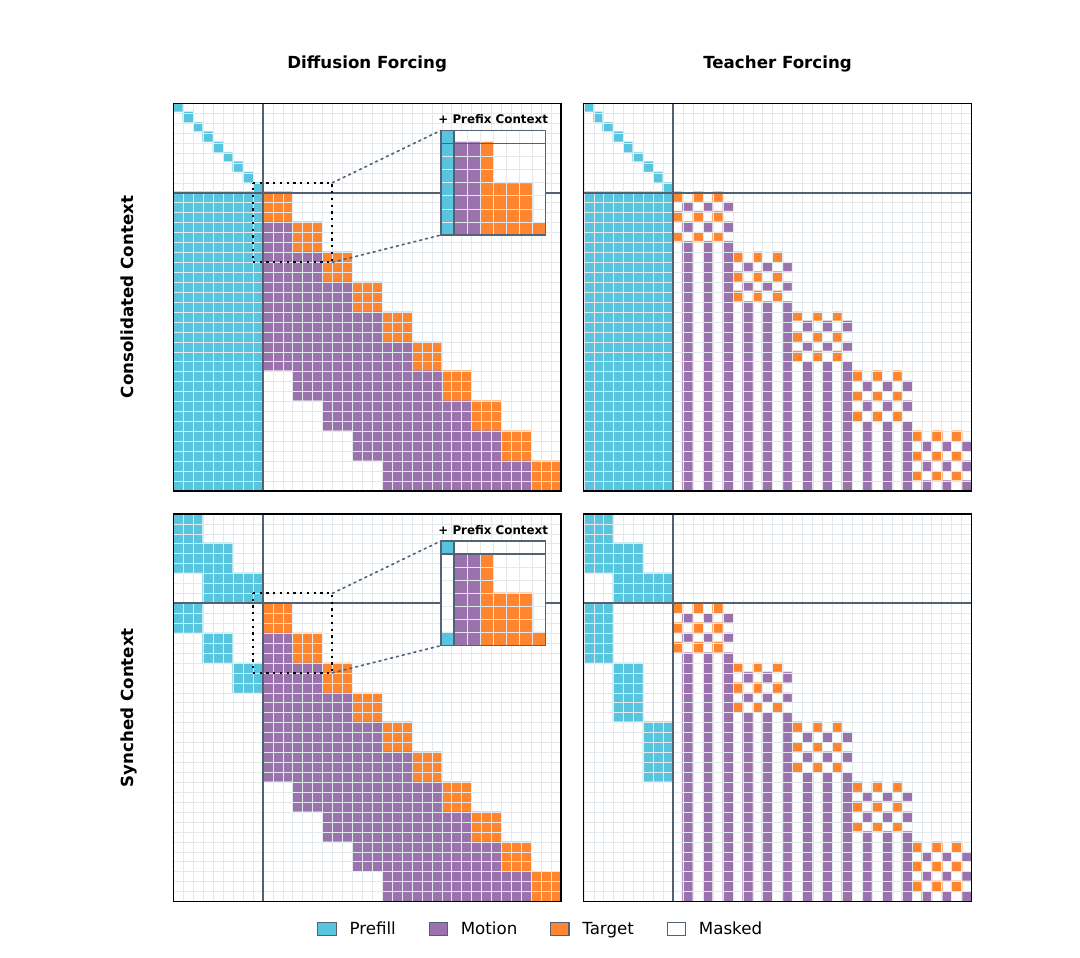}
\caption{\textbf{Sparse attention patterns.}
In each mask, rows index query views and columns index key views.
The first nine slots show prefill context.
The insets show diffusion forcing with \emph{prefix} context in the boxed regions.}%
\label{fig:supp-sparse-attention}
}
\paragraph{Sparse attention patterns.}
\figref{fig:supp-sparse-attention} illustrates sparse attention patterns for a target block size of $L^{\mathrm{tgt}}=3$.
Attention is bidirectional within blocks and causal across blocks.
Every target block can attend to \emph{consolidated} context from images or clips without timestamps.
\emph{Synced} context retains timestamps and aligns to target blocks, with a six-view context window.
TF masks attention between the clean and noisy copies of each block.
DF uses an 18-view window to exclude older target blocks and can also use clean \emph{prefix} context.

\newcommand{\AVMDistillationFigure}{%
\includegraphics[width=\linewidth]{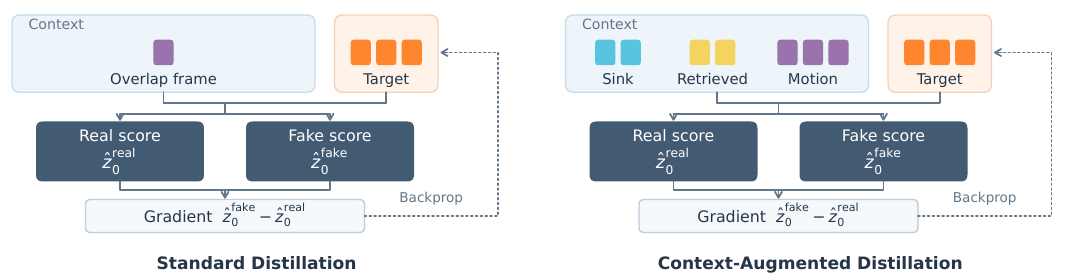}
\caption{\textbf{Standard distillation \emph{vs.}\ context-augmented distillation.}
Standard distillation (left) conditions both score models on one overlap frame.
Context-augmented distillation (right) uses \emph{sink}, \emph{retrieved}, and \emph{motion} context.
Both models score the same target block, and the difference between their denoised estimates drives the generator update.}%
\label{fig:supp-on-policy-distillation}
}
\ifAVMWrapFigures
\begin{figure}[H]
\centering
\AVMSparseAttentionFigure
\end{figure}
\fi

\paragraph{Distillation on rollouts of varying length.}
In standard distillation~\citep{yin2024onestep}, the frozen base model supplies the \emph{real} score through its denoised estimate $\hat{\mathbf z}_0^{\mathrm{real}}$.
A separate denoiser trained by flow matching on the generator's outputs supplies the \emph{fake} score through $\hat{\mathbf z}_0^{\mathrm{fake}}$.
Both denoise the same noisy target latents given the same context.
The normalized difference between their estimates, $\hat{\mathbf z}_0^{\mathrm{fake}}-\hat{\mathbf z}_0^{\mathrm{real}}$, supplies the generator gradient.
For base or causal rollouts of varying length, a \emph{scoring chunk} pairs a short generated target segment with selected context views.
Scoring and backpropagation stay local to the target latents in each chunk~\citep{yang2026longlive}.
To anchor the next chunk, the VAE decodes the previous chunk's overlap latent online and re-encodes its last RGB frame as a standalone image (left panel of \figref{fig:supp-on-policy-distillation}).

\paragraph{Context-augmented distillation.}
In context-augmented distillation, both score models instead condition each scoring chunk on a shared set of views from prefill and history.
Prefill views can persist as \emph{sink} context, and we select \emph{retrieved} views from prefill or history by camera pose or timestamp (right panel of \figref{fig:supp-on-policy-distillation}).
\emph{Motion} context can span more than one latent slot immediately before the target to preserve motion across chunk boundaries.
These sources extend the scoring context beyond the single overlap frame used in standard distillation.

\ifAVMWrapFigures
\begin{figure}[!ht]
\centering
\AVMDistillationFigure
\end{figure}
\fi

\paragraph{Camera trajectory augmentation.}
During on-policy distillation, we use two primary augmentation modes (\tabref{tab:camera_augmentation}):
1) \emph{Reflected.} We reflect the camera poses of short sequences in an out-and-back pattern to reach the requested length while advancing scene time.
2) \emph{Synthetic.} We sample a path from the camera primitives in \figref{fig:supp-camera-trajectory-augmentation}.
The path starts at the anchor pose with its intrinsics.
For distillation and evaluation on the Ws split, we scale translations and orbit radii by the anchor's distance from the median camera center after gravity alignment.

\begin{table}[!ht]
\centering
\caption{\textbf{Camera trajectory augmentation during on-policy tuning.}
Indented probabilities are conditional on selecting a synthetic path and are rounded.
Variants are sampled uniformly within each family.}%
\label{tab:camera_augmentation}
\small
\renewcommand{\arraystretch}{1.15}
\begin{tabularx}{\linewidth}{@{}lc>{\raggedright\arraybackslash}X@{}}
\toprule
\textbf{Path or family} & \textbf{Prob.\ (\%)} & \textbf{Motion variants} \\
\midrule
\textbf{Reflected} & \textbf{80.0} & \textbf{Reflect the original ground-truth trajectory to the requested length.} \\
\midrule
\textbf{Synthetic} & \textbf{20.0} & \textbf{Synthesize a trajectory of the requested length using the primitives below.} \\
\hspace{0.8em}Static & 1.7 & Fixed camera. \\
\hspace{0.8em}Translation & 56.5 & Dolly in \& back, truck left \& right, pedestal up. \\
\hspace{0.8em}Rotation & 13.0 & Pan left \& right, tilt up \& down, roll in either direction. \\
\hspace{0.8em}Zoom & 2.6 & Zoom in \& out. \\
\hspace{0.8em}Orbital & 13.0 & Orbit, orbit left \& right, spiral, arc, boomerang, figure-eight. \\
\hspace{0.8em}Revisit & 13.0 & Reflected translation, translation \& rotation, loop closure. \\
\bottomrule
\end{tabularx}
\end{table}

\ifAVMWrapFigures\else
\begin{figure*}[t]
\centering
\AVMSparseAttentionFigure
\end{figure*}
\begin{figure*}[t]
\centering
\AVMDistillationFigure
\end{figure*}
\fi

\section{Evaluation Protocol}%
\label{sec:appendix_eval}

\paragraph{Metric resolution.}
We rescale and center-crop generated frames and any corresponding reference frames to $448\times448$ before scoring.
This lets us compare models with different native output resolutions at the same scoring resolution.

\paragraph{Timestamp.}
To hold a scene static, we use a constant timestamp across frames, as in static evaluations on the D and Ws splits.
For dynamic scenes, such as those on the Wd split, timestamps advance across frames to represent scene motion.

\paragraph{D split.}
We use all 141 DL3DV-Benchmark scenes.

\paragraph{Ws split.}
For the \emph{translation \& rotation} trajectory (used to evaluate camera controllability and visual quality) and the \emph{loop closure} trajectory (used to evaluate revisits), we use the same 26 scenes.
For the original three-segment trajectory (used to evaluate text and object controllability), we use a separate set of 20 scenes.
Both subsets contain equal numbers of photorealistic and stylized scenes and are listed in \tabref{tab:evaluation_scenes}.

\paragraph{Wd split.}
For the original fixed-camera trajectory (used to evaluate dynamics), we use a 32-scene subset spanning all five motion categories (\tabref{tab:evaluation_scenes}).

\paragraph{MultiCam split.}
For 4D reconstruction, we use the scenes listed in \tabref{tab:evaluation_scenes}.
We use \texttt{camera~01} as input and score all 81 synchronized frames from \texttt{camera~02}.

\section{Evaluation Metrics}%
\label{sec:appendix_metrics}

\subsection{Controllability}

\paragraph{Rotation error (Rot.).}
Reconstruction methods (e.g., COLMAP~\citep{schonberger2016sfm}) can estimate camera poses from generated clips.
Following WorldScore~\citep{duan2025worldscore}, we measure the mean geodesic angle $\bar r$, in degrees, between the estimated and requested rotations.
Lower is better.

\paragraph{Relative translation error (Trans.).}
We adapt WorldScore's position error~\citep{duan2025worldscore} by dividing the mean aligned position error by twice the mean distance of the requested cameras from the starting pose.
Both errors are observable on \emph{translation \& rotation} paths.
Lower is better.

\paragraph{Camera control (Camera).}
We adapt WorldScore's geometric mean~\citep{duan2025worldscore} to combine rotation error $\bar r$ and relative translation error $\bar t$ as $100\sqrt{(1-\operatorname{clip}(\bar r/15,0,1)) (1-\operatorname{clip}(\bar t/0.5,0,1))}$.
Higher is better.

\paragraph{Object controllability (Obj.).}
We use WorldScore's open-set object-detection success score~\citep{duan2025worldscore} at frames 0, 80, 160, and 240.
TransNet V2~\citep{soucek2020transnetv2} flags the first cut or gradual transition whose probability exceeds $0.5$; normalized anchor scores at or after that transition are set to zero before all four anchors are averaged.
Clips without a detected transition retain WorldScore's original aggregation.
Higher is better.

\paragraph{Text alignment (Text).}
We use WorldScore's text alignment~\citep{duan2025worldscore}, which compares each anchor frame with the full scene description using CLIP~\citep{radford2021clip}.
The same four anchors and TransNet V2~\citep{soucek2020transnetv2} transition gate used for object controllability prevent a scene change from improving text alignment at the expense of continuity.
Higher is better.

\subsection{Visual Quality}

\paragraph{Aesthetic quality (Aesth.).}
Following VBench~\citep{huang2024vbench}, the LAION aesthetic predictor~\citep{laion2022aesthetic} scores every frame on a 0--10 scale; we divide by ten and average over all evaluated frames and scenes.
Higher is better.

\paragraph{Imaging quality (Image).}
Following VBench~\citep{huang2024vbench}, MUSIQ~\citep{ke2021musiq} predicts the perceptual image quality of every frame, including the effects of blur, noise, and exposure.
We average the raw MUSIQ scores over all evaluated frames and scenes.
Higher is better.

\paragraph{Fr\'echet Inception distance (FID).}
We pool Inception features~\citep{szegedy2016inception} from all evaluated generated frames and their reference frames, and compute one global FID~\citep{heusel2017fid} for each method and dataset.
Lower is better.

\paragraph{Video multi-method assessment fusion (VMAF).}
VMAF~\citep{netflix2016vmaf} computes a full-reference perceptual video score between each generated sequence and its aligned target sequence.
We average sequence scores over scenes.
Higher is better.

\paragraph{DreamSim distance (DSim).}
DreamSim~\citep{fu2023dreamsim} measures the learned perceptual distance between each generated frame and its corresponding target frame.
We average over evaluated frames and scenes.
Lower is better.

\subsection{Dynamics}

\paragraph{Motion accuracy (Acc.).}
WorldScore motion accuracy~\citep{duan2025worldscore} compares optical-flow magnitude inside the prompt-designated moving region with flow outside it, discounting global motion caused by camera movement.
Higher is better.

\paragraph{Dynamic degree (Deg.).}
VBench~\citep{huang2024vbench} uses RAFT~\citep{teed2020raft} to estimate optical flow between adjacent frames, averages the largest 5\% of flow magnitudes, and classifies each video as static or dynamic.
Dynamic degree is the proportion of videos classified as dynamic.
Higher is better.

\paragraph{Motion smoothness (Smooth.).}
VBench~\citep{huang2024vbench} removes odd-numbered frames, reconstructs them with the AMT frame interpolation model~\citep{li2023amt}, and converts the reconstruction error to a normalized smoothness score.
Higher is better.

\subsection{Scene Persistence}

\paragraph{Peak signal-to-noise ratio (PSNR).}
PSNR~\citep{ling2024dl3dv} compares corresponding generated and target RGB frames for input-context reconstruction, or, for generated-history revisits, the two frames generated at the revisited pose.
Scores are averaged within each scene and then across scenes.
Higher is better.

\paragraph{Structural similarity (SSIM).}
SSIM~\citep{wang2004ssim} compares luminance, contrast, and local structure for the same frame pairs used by PSNR\@.
Scores are averaged within each scene and then across scenes.
Higher is better.

\paragraph{Learned perceptual image patch similarity (LPIPS).}
LPIPS~\citep{zhang2018lpips} measures deep-feature distance for the same frame pairs.
Scores are averaged within each scene and then across scenes.
Lower is better.

\paragraph{Matching pixels (Mat.\ Pix.).}
We use GIM~\citep{shen2024gim} to match pixels between corresponding generated and target frames and count correspondences above the benchmark confidence threshold.
We average counts over frames and scenes and report the result in thousands.
Higher is better.

\paragraph{Place recognition (VPR).}
Following DreamX-World~\citep{dreamx2026world}, we extract one global MutualVPR descriptor~\citep{gu2025mutualvpr} from each image and report the cosine similarity between paired views.
This metric tests whether two views remain recognizable as the same place despite a viewpoint change.
Higher is better.

\paragraph{Geometric matching (SP).}
Following DreamX-World~\citep{dreamx2026world}, we detect up to 1,024 SuperPoint keypoints~\citep{detone2018superpoint} in each image and match them with LightGlue~\citep{lindenberger2023lightglue}.
For a pair with $N_i$ and $N_j$ detected keypoints and $N_{\mathrm{match}}$ accepted matches, we report $N_{\mathrm{match}}/\min(N_i,N_j)$.
The score is undefined when either image has no detected keypoints, which occurs for six HY-WorldPlay revisit pairs.
Higher is better.

\paragraph{3D consistency (3D Cons.).}
Following WorldScore~\citep{duan2025worldscore}, DROID-SLAM~\citep{teed2021droidslam} estimates dense depth and camera motion, and the metric measures reprojection error between covisible pixels in consecutive frames.
The benchmark converts this error to a normalized consistency score.
Higher is better.

\section{Context Retrieval with Camera Frustums}%
\label{sec:appendix_camera_retrieval}

\begin{figure*}[!t]
\centering
\includegraphics[width=\linewidth]{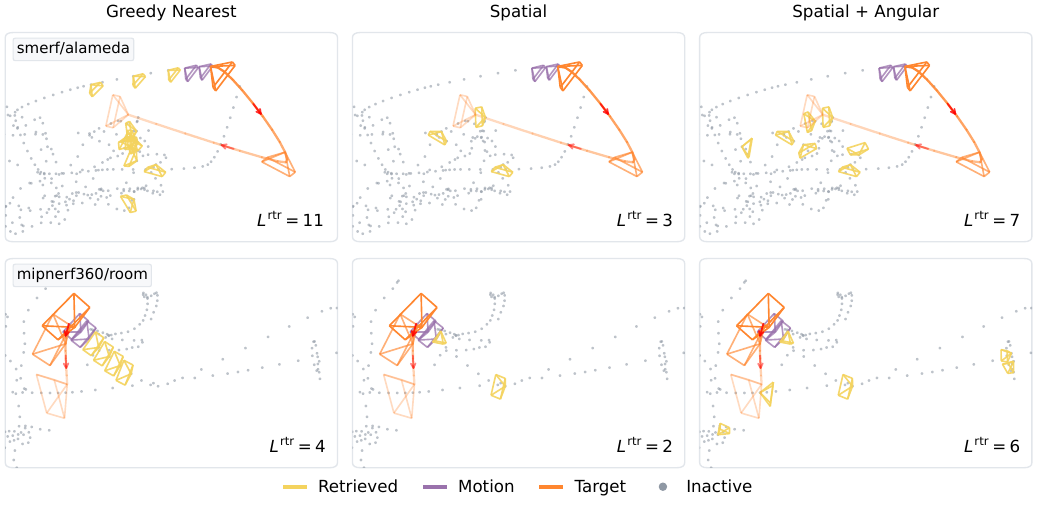}
\caption{\textbf{Comparison of camera-based retrieval policies.}
We show the fourth chunk from \texttt{smerf/alameda}~\citep{duckworth2024smerf} (top) and \texttt{mipnerf360/room}~\citep{barron2022mipnerf360} (bottom).
All panels in a row share an orthographic view and scale.
$L^{\mathrm{rtr}}$ is the retrieved context length (see \secref{sec:sampling}).}%
\label{fig:camera-retrieval-ablation}
\end{figure*}

\paragraph{Retrieval policies.}
Let $\mathcal{R}$ be the pool of prefill and history views excluding \emph{sink} and \emph{motion} context.
Retriever size $s$ sets the number of slots budgeted per target, giving a total budget of $\min(sL^{\mathrm{tgt}},|\mathcal{R}|)$ per chunk.
We compare three retrieval policies:
1) \emph{Greedy nearest.} For each target query $q$, rank views $r$ by \mbox{$d(q,r)=\lVert P_q^{-1}(P_r-P_q)\rVert_{\mathrm{F}}$}, where $P_q$ and $P_r$ are $4{\times}4$ camera projection matrices with normalized image coordinates and $\lVert\cdot\rVert_{\mathrm{F}}$ is the Frobenius norm.
Select the $s$ nearest views per target and deduplicate by view index.
2) \emph{Spatial.} Sample 3D anchors along rays through a regular $g{\times}g$ image grid for each target.
Anchor depths are fixed multiples of the median nonzero camera-center distance from the target to views in $\mathcal{R}$.
Within a radius proportional to the camera-center distance from the target to its $k$-th nearest neighbor in $\mathcal{R}$, a candidate covers anchors that project inside its image at positive depth.
Repeatedly retrieve the candidate covering the most uncovered anchors across all targets, updating coverage after each retrieval.
Stop when the budget is full or no candidate adds coverage.
3)~\mbox{\textbf{\emph{Spatial + angular}} (default).} In addition to the spatial pass, retrieve up to four views to fill missing target-relative headings (i.e., parallel, opposite, left, or right) and maintain viewpoint diversity as targets turn.
The angular pass can search beyond the spatial radius, prioritizing views that overlap the target.
Both coverage policies apply pose non-maximum suppression (NMS) after all passes without refilling slots.

\ifAVMWrapFigures
\Needspace{10\baselineskip}
\begin{wraptable}{r}{.48\linewidth}
\else
\begin{table}[!ht]
\fi
\centering
\caption{\textbf{Coverage and context length by retrieval policy.}
We average over chunks within each scene, then weight scenes equally.
$L^{\mathrm{rtr}}$ counts retrieved slots per chunk.
Empty is the percentage of chunks with no retrieved views.
Slot counts and overlap exclude \emph{motion} context.
Bold marks the default policy.}%
\label{tab:camera-retrieval-policy}
\ifAVMWrapFigures\footnotesize\else\small\fi
\renewcommand{\arraystretch}{1.15}
\setlength{\tabcolsep}{4pt}
\begin{tabularx}{\linewidth}{@{}>{\raggedright\arraybackslash}Xrrr}
\toprule
\textbf{Retrieval policy} & Mean $L^{\mathrm{rtr}}\uparrow$ & Overlap (\%)$\uparrow$ & Empty (\%)$\downarrow$ \\
\midrule
Greedy nearest & \besttwo{5.3} & \besttwo{73.0} & \best{0.0} \\
Spatial & \bestthree{3.5} & \bestthree{65.1} & \besttwo{22.0} \\
\textbf{Spatial + angular} & \best{6.5} & \best{84.4} & \best{0.0} \\
\bottomrule
\end{tabularx}
\ifAVMWrapFigures
\end{wraptable}
\else
\end{table}
\fi

\paragraph{Policy comparison.}
We evaluate the three retrieval policies on selected scenes by replaying camera poses under the base model's chunk-wise autoregressive schedule, using $L^{\mathrm{win}}=21$ throughout and $L^{\mathrm{mot}}=2$ from the second chunk onward.
At $s=1$, we report three metrics.
Mean $L^{\mathrm{rtr}}$ is the average number of retrieved slots per chunk.
Overlap (\%) is the share of samples on a separate target grid covered by retrieved views.
Empty (\%) is the share of chunks without retrieved views.
As shown in \tabref{tab:camera-retrieval-policy}, \emph{spatial + angular} leads in mean overlap (84.4\% \emph{vs.}\ 73.0\% for \emph{greedy nearest} and 65.1\% for \emph{spatial}), retrieves more slots on average, and eliminates the empty chunks seen with \emph{spatial}.
\figref{fig:camera-retrieval-ablation} illustrates how the default policy retrieves more diverse viewpoints while preserving the coverage of the spatial pass.

\begin{table*}[!t]
\centering
\caption{\textbf{Selected evaluation scenes.}
WorldScore and MultiCam scene identifiers for the settings in \secref{sec:appendix_eval}.}%
\label{tab:evaluation_scenes}
\renewcommand{\ttdefault}{lmtt}
\fontsize{7.3}{9}\selectfont
\setlength{\tabcolsep}{0pt}
\begin{tabular*}{\linewidth}{@{\extracolsep{\fill}}ll@{}}
\toprule
\multicolumn{2}{@{}l}{\textbf{Ws split (26 scenes)} for the \emph{translation \& rotation} and \emph{loop closure} trajectories} \\
\verb|static_photorealistic_indoor_dining_spaces_002_000002| & \verb|static_photorealistic_indoor_dining_spaces_006_000006| \\
\verb|static_photorealistic_indoor_dining_spaces_028_000028| & \verb|static_photorealistic_indoor_living_spaces_019_000119| \\
\verb|static_photorealistic_indoor_passageways_002_000202| & \verb|static_photorealistic_indoor_passageways_011_000211| \\
\verb|static_photorealistic_indoor_passageways_013_000213| & \verb|static_photorealistic_indoor_passageways_044_000244| \\
\verb|static_photorealistic_indoor_public_spaces_061_000361| & \verb|static_photorealistic_outdoor_aquatic_landscape_006_000506| \\
\verb|static_photorealistic_outdoor_city_023_000623| & \verb|static_photorealistic_outdoor_suburb_041_000741| \\
\verb|static_photorealistic_outdoor_terrestrial_landscape_056_000856| & \verb|static_stylized_indoor_dining_spaces_006_001006| \\
\verb|static_stylized_indoor_dining_spaces_012_001012| & \verb|static_stylized_indoor_dining_spaces_043_001043| \\
\verb|static_stylized_indoor_dining_spaces_045_001045| & \verb|static_stylized_indoor_dining_spaces_050_001050| \\
\verb|static_stylized_indoor_living_spaces_067_001167| & \verb|static_stylized_indoor_public_spaces_002_001302| \\
\verb|static_stylized_indoor_public_spaces_012_001312| & \verb|static_stylized_indoor_workspaces_029_001429| \\
\verb|static_stylized_outdoor_aquatic_landscape_054_001554| & \verb|static_stylized_outdoor_city_071_001671| \\
\verb|static_stylized_outdoor_terrestrial_landscape_009_001809| & \verb|static_stylized_outdoor_verdant_landscape_024_001924| \\
\midrule
\multicolumn{2}{@{}l}{\textbf{Ws split (20 scenes)} for the original three-segment trajectory} \\
\verb|static_photorealistic_indoor_dining_spaces_080_000080| & \verb|static_photorealistic_indoor_workspaces_089_000489| \\
\verb|static_stylized_outdoor_aquatic_landscape_086_001586| & \verb|static_photorealistic_indoor_dining_spaces_081_000081| \\
\verb|static_photorealistic_outdoor_terrestrial_landscape_088_000888| & \verb|static_photorealistic_outdoor_city_081_000681| \\
\verb|static_stylized_outdoor_terrestrial_landscape_096_001896| & \verb|static_stylized_indoor_passageways_088_001288| \\
\verb|static_stylized_outdoor_verdant_landscape_091_001991| & \verb|static_photorealistic_outdoor_aquatic_landscape_084_000584| \\
\verb|static_photorealistic_indoor_dining_spaces_082_000082| & \verb|static_photorealistic_indoor_public_spaces_086_000386| \\
\verb|static_photorealistic_outdoor_city_094_000694| & \verb|static_photorealistic_outdoor_verdant_landscape_099_000999| \\
\verb|static_stylized_indoor_dining_spaces_080_001080| & \verb|static_stylized_indoor_living_spaces_099_001199| \\
\verb|static_stylized_indoor_public_spaces_099_001399| & \verb|static_stylized_outdoor_aquatic_landscape_099_001599| \\
\verb|static_stylized_outdoor_suburb_098_001798| & \verb|static_stylized_outdoor_verdant_landscape_099_001999| \\
\midrule
\multicolumn{2}{@{}l}{\textbf{Wd split (32 scenes)} for the original fixed-camera trajectory} \\
\verb|dynamic_photorealistic_articulated_articulated_000_000099| & \verb|dynamic_photorealistic_articulated_articulated_015_000014| \\
\verb|dynamic_photorealistic_articulated_articulated_017_000016| & \verb|dynamic_photorealistic_articulated_articulated_020_000019| \\
\verb|dynamic_photorealistic_articulated_articulated_068_000067| & \verb|dynamic_photorealistic_articulated_articulated_085_000084| \\
\verb|dynamic_photorealistic_articulated_articulated_098_000097| & \verb|dynamic_photorealistic_deformable_deformable_000_000199| \\
\verb|dynamic_photorealistic_deformable_deformable_012_000111| & \verb|dynamic_photorealistic_deformable_deformable_035_000134| \\
\verb|dynamic_photorealistic_deformable_deformable_040_000139| & \verb|dynamic_photorealistic_deformable_deformable_047_000146| \\
\verb|dynamic_photorealistic_deformable_deformable_083_000182| & \verb|dynamic_photorealistic_deformable_deformable_084_000183| \\
\verb|dynamic_photorealistic_fluid_fluid_016_000215| & \verb|dynamic_photorealistic_fluid_fluid_024_000223| \\
\verb|dynamic_photorealistic_fluid_fluid_042_000241| & \verb|dynamic_photorealistic_fluid_fluid_053_000252| \\
\verb|dynamic_photorealistic_fluid_fluid_078_000277| & \verb|dynamic_photorealistic_fluid_fluid_080_000279| \\
\verb|dynamic_photorealistic_multi_motion_multi_motion_001_000300| & \verb|dynamic_photorealistic_multi_motion_multi_motion_028_000327| \\
\verb|dynamic_photorealistic_multi_motion_multi_motion_076_000375| & \verb|dynamic_photorealistic_multi_motion_multi_motion_094_000393| \\
\verb|dynamic_photorealistic_multi_motion_multi_motion_095_000394| & \verb|dynamic_photorealistic_rigid_rigid_000_000499| \\
\verb|dynamic_photorealistic_rigid_rigid_006_000405| & \verb|dynamic_photorealistic_rigid_rigid_027_000426| \\
\verb|dynamic_photorealistic_rigid_rigid_046_000445| & \verb|dynamic_photorealistic_rigid_rigid_073_000472| \\
\verb|dynamic_stylized_multi_motion_multi_motion_020_000820| & \verb|dynamic_stylized_rigid_rigid_060_000960| \\
\midrule
\multicolumn{2}{@{}l}{\textbf{MultiCam split (24 held-out scenes)} for the original trajectory} \\
\multicolumn{2}{@{}l}{18 and 24~mm: \texttt{1}, \texttt{616}, \texttt{1764}, \texttt{2146}, \texttt{2529}, \texttt{2911}, \texttt{3294}} \\
\multicolumn{2}{@{}l}{35 and 50~mm: \texttt{1}, \texttt{616}, \texttt{1764}, \texttt{2529}, \texttt{2911}} \\
\bottomrule
\end{tabular*}
\end{table*}

\begin{samepage}
\section{Additional Qualitative Results}%
\label{sec:appendix_capabilities}

We expand on the capabilities shown in \figref{fig:full-page} with additional examples.
Each row shows representative context views with their recorded camera frustums (blue), the requested camera trajectory (dotted orange), and three target views with their frustums (orange).
Video prompts show sampled frames; dense image prompts show a subset of the supplied images.
All panels within a row use the same source run, retaining its native images and camera calibration.

\end{samepage}

\clearpage

\begin{figure*}[p]
\centering
\includegraphics[width=\textwidth]{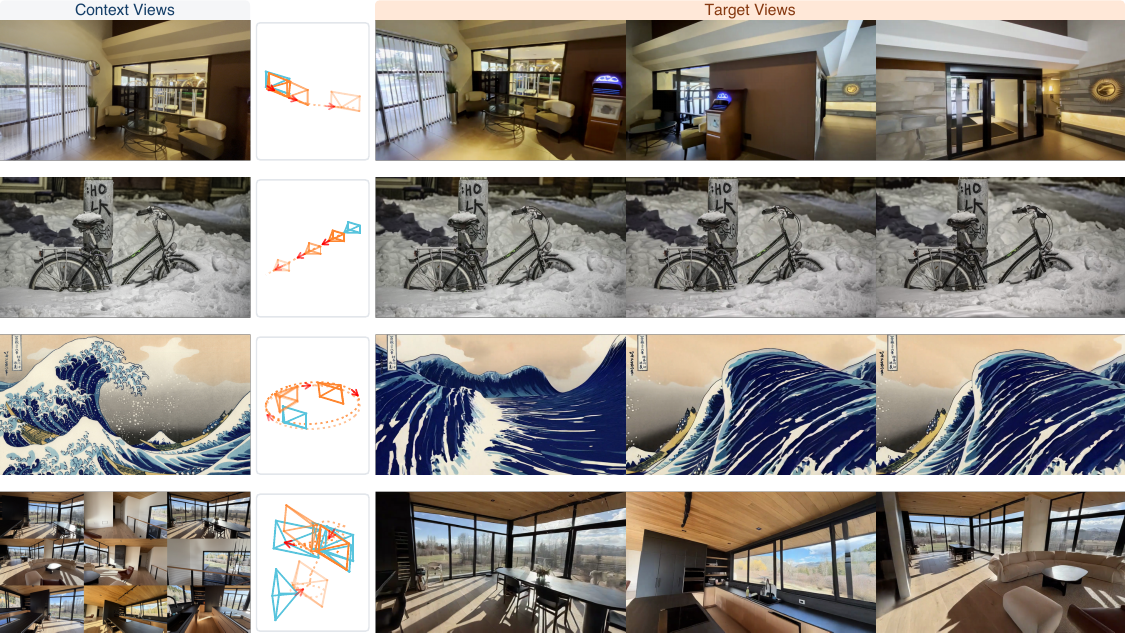}
\caption{\textbf{Camera control.}
AVM uses image context to generate target views along requested camera paths in photographic and stylized scenes.}%
\label{fig:supp-camera-control}
\end{figure*}

\begin{figure*}[p]
\centering
\includegraphics[width=\textwidth]{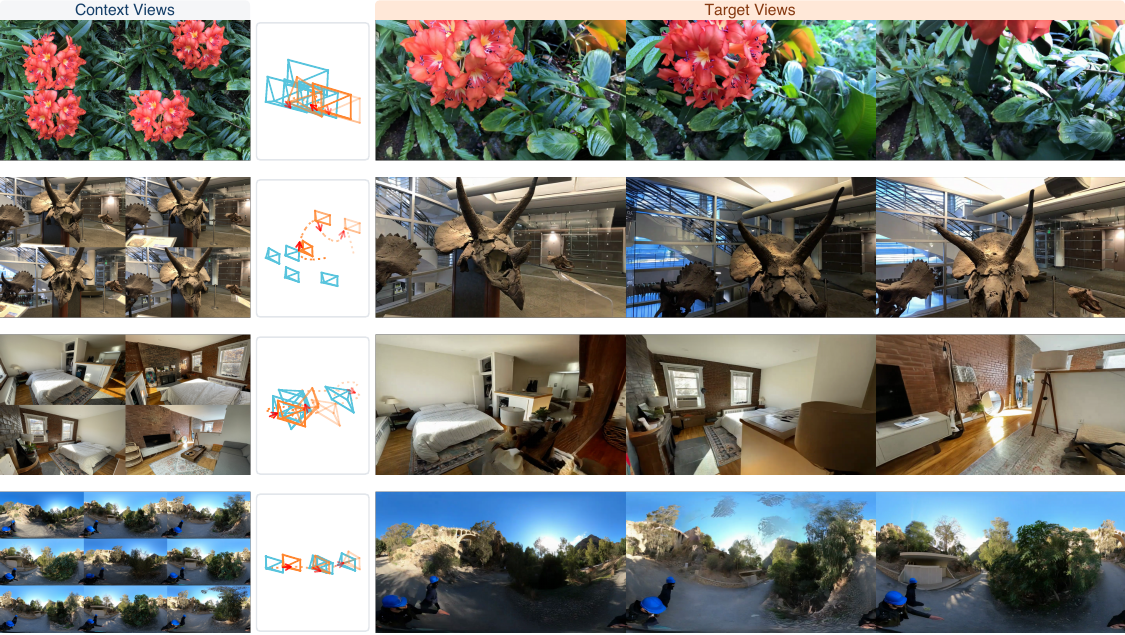}
\caption{\textbf{Image context.}
One or more posed images provide context for target views at the requested camera poses.\AVMFaceBlurNotice}%
\label{fig:supp-image-context}
\end{figure*}

\begin{figure*}[p]
\centering
\includegraphics[width=\textwidth]{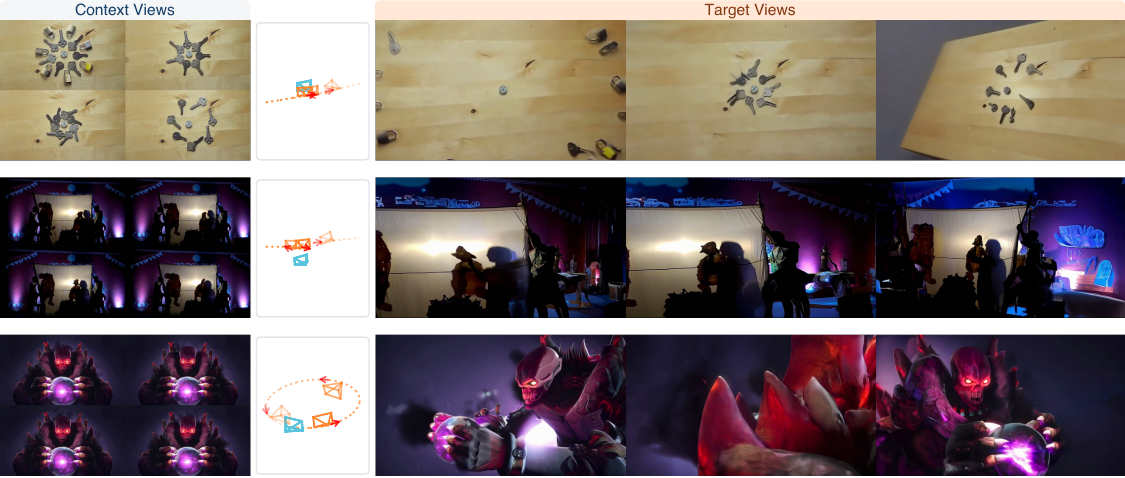}
\caption{\textbf{Video stream context.}
A single input video provides context for target views at requested camera poses and timestamps.
Each mosaic shows sampled input frames.}%
\label{fig:supp-video-stream-context}
\end{figure*}

\begin{figure*}[p]
\centering
\includegraphics[width=\textwidth]{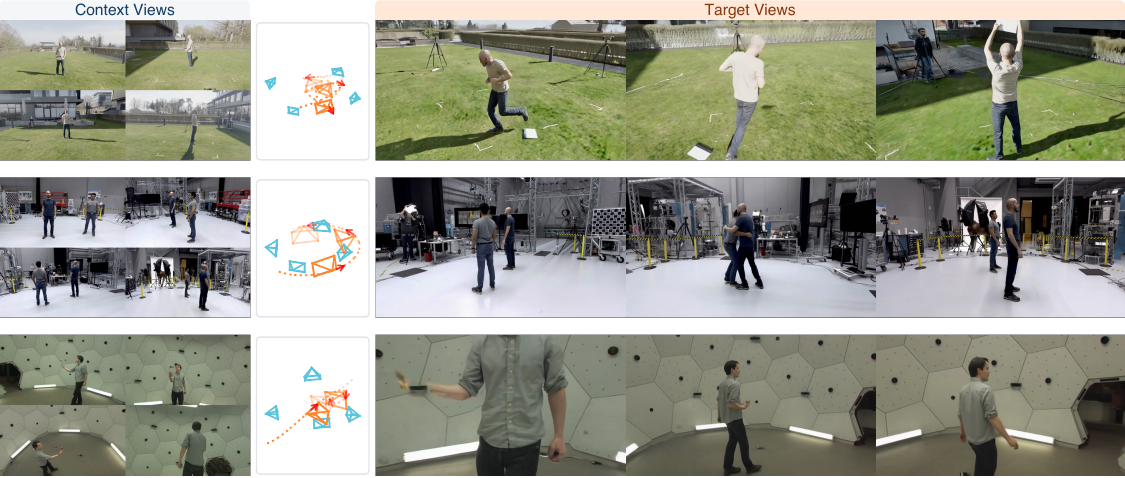}
\caption{\textbf{Video rig context.}
Synchronized streams from four camera rigs provide \emph{synced} context at the target timestamps.\AVMFaceBlurNotice}%
\label{fig:supp-video-rigs-context}
\end{figure*}

\begin{figure*}[p]
\centering
\includegraphics[width=\textwidth]{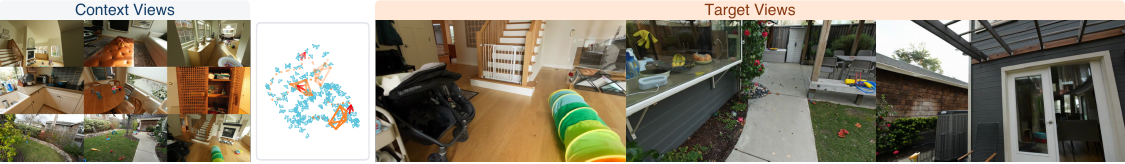}
\caption{\textbf{Dense view prefilling.}
A dense set of posed images provides prefill context for target views along the requested trajectory.
The mosaic shows a subset of the inputs.}%
\label{fig:supp-dense-view-prefilling}
\end{figure*}

\begin{figure*}[p]
\centering
\includegraphics[width=\textwidth]{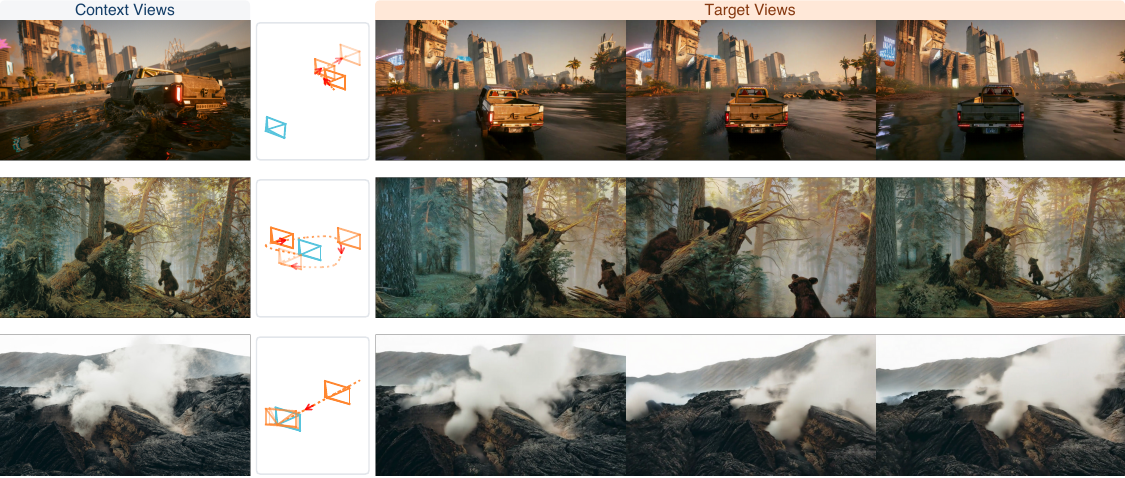}
\caption{\textbf{Dynamics.}
Target views show scene motion as timestamps advance and the camera follows the requested trajectory.}%
\label{fig:supp-dynamics}
\end{figure*}

\begin{figure*}[p]
\centering
\includegraphics[width=\textwidth]{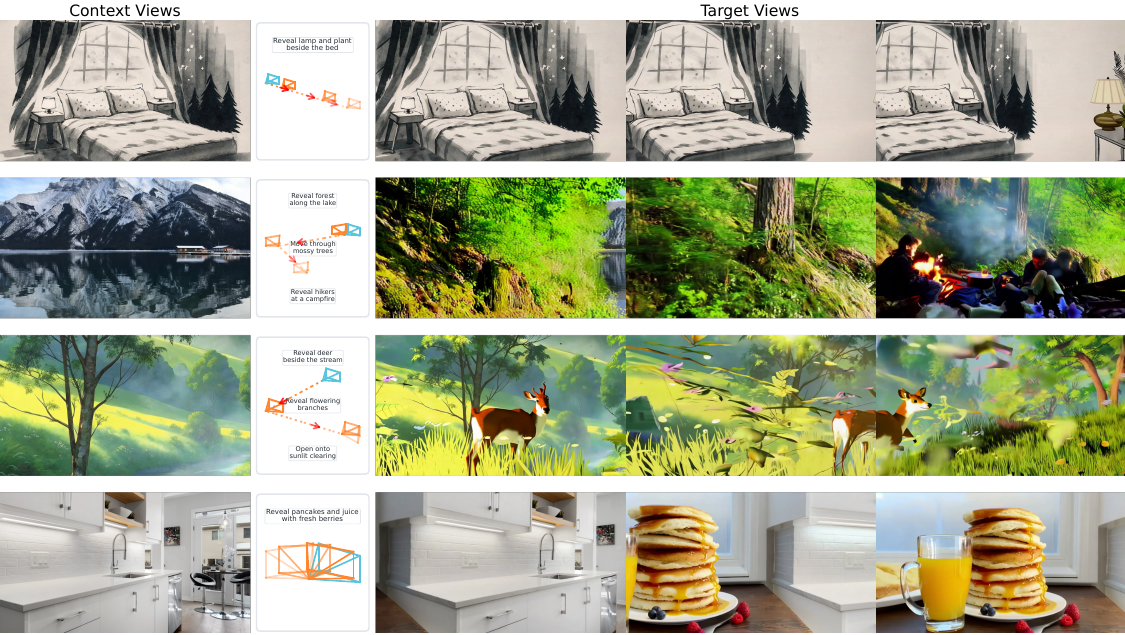}
\caption{\textbf{Event prompting.}
Instruction labels specify the requested content changes as AVM generates target views along the camera path.}%
\label{fig:supp-event-prompting}
\end{figure*}

\begin{figure*}[p]
\centering
\includegraphics[width=\textwidth]{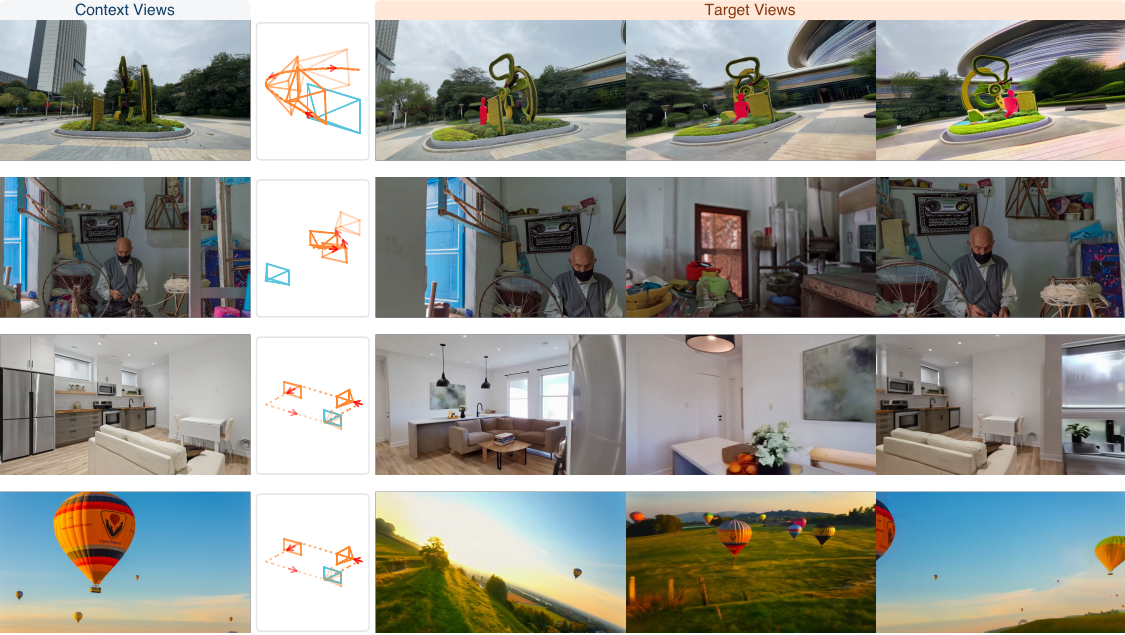}
\caption{\textbf{Long-horizon consistency.}
Target views at revisited locations illustrate scene persistence after long camera trajectories in static and dynamic scenes.\AVMFaceBlurNotice}%
\label{fig:supp-long-horizon-consistency}
\end{figure*}

\begin{figure*}[p]
\centering
\includegraphics[width=\textwidth]{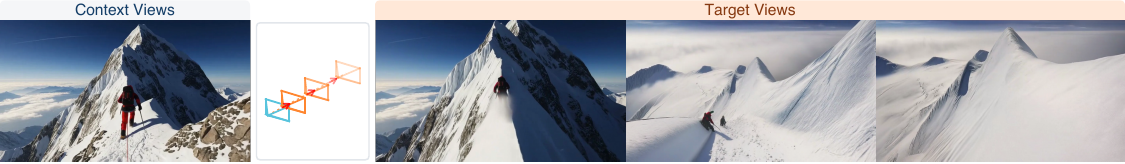}
\caption{\textbf{Perpetual view streaming.}
AVM generates successive blocks along an extended camera path.
New target views extend the stream and become available as history context.}%
\label{fig:supp-perpetual-view-streaming}
\end{figure*}

\end{document}